\documentclass{article}

\usepackage{arxiv}

\usepackage[utf8]{inputenc} 
\usepackage{algorithm}\usepackage{algorithmic}
\usepackage{booktabs}
\usepackage{wrapfig}
\usepackage{paralist}
\usepackage{graphicx}
\newcommand\widebar[1]{\mathop{\overline{#1}}}
\usepackage{microtype}
\usepackage{graphicx}
\usepackage{subfigure}
\usepackage{booktabs} 
\usepackage{multirow}
\usepackage{multicol}
\usepackage{array}
\usepackage{amsmath}
\usepackage{amsfonts}
\usepackage{mathrsfs}
\usepackage{graphics,wrapfig} 
\usepackage{paralist}
\usepackage{comment}
\usepackage{placeins}

\title{Scaling Bayesian Optimization with Game Theory}

\author{
 Logan Mathesen \\
  School of Computing and Augemented Intelligence\\
  Arizona State University\\
  Tempe, AZ 85258 \\
  \texttt{lmathese@asu.edu} \\
   \And
  Giulia Pedrielli \\
  School of Computing and Augemented Intelligence\\
  Arizona State University\\
  Tempe, AZ 85258 \\
  \texttt{gpedriel@asu.edu} \\
   \AND
   Robert L. Smith \\
   Industrial and Operations Engineering \\
   University of Michigan \\
   Ann Arbor, MI 48109 \\
   \texttt{rlsmith@umich.edu} \\
}

\date{}

\begin{document}

\maketitle


\begin{abstract}
    We introduce the algorithm Bayesian Optimization with Fictitious Play (BOFiP) for the optimization of high dimensional black box functions. BOFiP decomposes the original, high dimensional, space into several sub-spaces defined by non-overlapping sets of dimensions. These sets are randomly generated at the start of the algorithm, and they form a partition of the dimensions of the original space. BOFiP can search the original space through a strategic learning mechanism that alternates Bayesian optimization, to search within sub-spaces, and information exchange among sub-spaces, to update the sub-space function evaluation. The basic idea is to achieve increased efficiency by distributing the high dimensional optimization across low dimensional sub-spaces. In particular, each sub-space is interpreted as a player within an equal interest game. At each iteration, Bayesian optimization produces approximate best replies that allow the update of the players belief distribution. The belief update and Bayesian optimization continue to alternate until a stopping condition is met.

High dimensional problems are often encountered in real applications, and several contributions in the Bayesian optimization literature have highlighted the difficulty in scaling to high dimensions due to the computational complexity associated to the estimation of the model hyperparameters. Such complexity is exponential in the problem dimension, resulting in substantial loss of performance for most techniques with the increase of the input dimensionality. 

We compare BOFiP to several state-of-the-art approaches in the field of high dimensional black box optimization. The numerical experiments show the performance over three benchmark objective functions from 20 up to 1000 dimensions. A neural network architecture design problem is tested with 42 up to 911 nodes in 6 up to 92 layers, respectively, resulting into networks with 500 up to 10,000 weights. These sets of experiments empirically show that BOFiP outperforms its competitors, showing consistent performance across different problems and increasing problem dimensionality.
\end{abstract}%




%


\section{Introduction}\label{sec::intro}
We consider the problem of global optimization of a high dimensional, black box, possibly non-linear non-convex, function. 
A black box function can only be evaluated at selected locations, and no gradient information is returned upon evaluation~\cite{regis2007stochastic,muller2019surrogate}. 
Such optimization problems are attracting interest in engineering (AI-controlled systems, certification of embedded systems), as well as scientific (scientific machine learning for emulation of highly complex physical phenomena) applications~\cite{bergstra2012random,sarafian2020explicit}.

Direct search methods as well as model based methods have been developed to solve black box optimization problems. Direct search methods, such as gradient based approaches~\cite{Spall2003,kolda2003optimization,amaran2016simulation}, or simulated annealing~\cite{atkinson1992segmented,bertsimas1993simulated}, utilize local information derived from sampling to move across the space (deterministically or probabilistically), but do not require the construction of a model of the function to take a sampling decision. 

On the other hand, model based approaches require the estimation of a surrogate (model) of the original function to iteratively select sampling locations~\cite{regis2007stochastic,regis2009parallel,negoescu2011knowledge,fan2018surrogate,muller2019surrogate,muller2017socemo}.  
Among surrogate-based approaches, Bayesian optimization has been quite successful~\cite{frazier2018tutorial, movckus1975bayesian,brochu2010tutorial,shahriari2016taking}, with several applications such as machine learning, automatic algorithm design~\cite{snoek2012practical,bergstra2011algorithms,mahendran2012adaptive}, and robotics~\cite{lizotte2008practical,martinez2009bayesian}. 

Despite the satisfactory empirical performance and vast applicability shown by the state-of-the-art approaches, Bayesian  optimization methods face substantial challenges when 
dealing with large data sets, and with high dimensional inputs. 

In its most common implementation, Bayesian optimization uses a Gaussian process as a surrogate of the unknown objective function~\cite{wang2013bayesian,nayebi2019framework,kandasamy2015high,rolland2018high,frazier2018tutorial,snoek2012practical}, and 
three main issues arise: 
\begin{inparaenum}\item[(1)] the hyperparameters of the likelihood function to be maximized form a ($d+1$) dimensional vector (with $d$ being the dimensionality of the input). Hence, the higher the dimensionality, the more time the Gaussian process estimation will take; 
\item[(2)] the sampling problem becomes more complex due to the increased difficulty of optimizing the sampling criteria; \item[(3)] the computational complexity of the Gaussian process estimation task is cubic in the number of sampled locations. 
While the last challenge also occurs in low dimensional models, it exacerbates in high dimensional cases due to the need of a larger number of points to achieve desired levels of prediction accuracy\end{inparaenum}. 
In this work, we directly address challenges $(1)$ and $(2)$ by controlling the dimensionality of all the inputs associated to the surrogates used during the search. Furthermore, given a number of function evaluations, BOFiP distributes them across several sub-problems, mitigating the effect of large sample sizes (challenge $(3)$).  

In the literature, several approaches have been proposed to scale Bayesian optimization. Projection based methods perform the search in lower dimensional manifolds, assuming a low effective dimensional space exists that contains all the behavioral information of the original, high dimensional, function~\cite{wang2013bayesian}. 
Additionally, there have been developments in statistical learning to scale Bayesian optimization. An effort has been dedicated to the design of new 
surrogate model forms that exploit structural properties of the original high dimensional function, such as additivity, resulting in substantial computational gains in model estimation as well as sampling decisions~\cite{kandasamy2015high}. 

Our approach for scaling Bayesian optimization to higher dimensions makes no assumptions on the existence of a low dimensional structure in the original problem. Also, our approach can help scaling projection based methods, especially when the effective dimension is still high. 
Our algorithm, Bayesian optimization with sampled fictitious play (BOFiP), 
leverages the theory of equal interest games to decompose the original high dimensional problem into lower dimensional, intelligently interconnected, problems~\cite{brown1951iterative,young2004strategic,swenson2017regular,swenson2017exponential,swenson2018best,lambert2005fictitious}. 
In particular, we decompose the high dimensional space into non-overlapping low dimensional sub-spaces (i.e., spaces defined by a subset of coordinates), such that each coordinate is present in exactly one sub-space. Then, we interpret each sub-space as a \textit{player} in an equal interest game. As a result, the sub-spaces communicate to one another, as players in a game, by sharing their ``local'' optimization results. Based on the local optimization results each sub-space updates its belief on the distribution of promising locations. 

\section{Related Work}\label{sec::SoTA}
    High dimensional optimization has been identified as an important and open research area in both discrete~\cite{xu2013adaptive,powell2010merging}, and continuous optimization~\cite{xu2016statistical}. 
    In continuous settings, we distinguish two main families of approaches for high dimensional black box optimization: \begin{inparaenum}
    \item[(i)] projection based; \item[(ii)] high dimensional statistical learning based\end{inparaenum}. Methods in the first class assume the existence of a low dimensional manifold sufficient to represent the original function. Algorithms in this family learn a lower dimensional manifold through projection mechanisms, and perform the search in the implied lower dimensional space. Approaches from the high dimensional statistics literature, instead, focus on the design of new statistical models and associated efficient learning algorithms. Rather than assuming the existence of low dimensional embeddings these methods exploit structural properties of the function being learned, such as additivity. 
    As a result, the search is performed in the original space and sampling decisions are made by maximizing acquisition functions based on the predictors associated with the statistical models. Sections~\ref{sec::sotaProj}-\ref{sec::sotahdsl} review these two families of approaches.
    
    
    \subsection{Projection based methods}\label{sec::sotaProj}
        Projection based approaches tackle high dimensional optimization by exploiting
        the concept of effective dimensionality. A function 
        is said to have effective dimensionality 
        if there exists a set of linearly independent vectors that implies a space of dimensionality lower than the original space, such that any point in the original space can be projected onto the implied, lower dimensional, space without any change in the function value~\cite{dalalyan2008new}. The dimensionality of the implied space is the effective dimensionality of the original problem. Effective dimensionality based approaches will perform well in high dimensional settings when the effective dimensionality is much lower than the original: in this way both the model estimation task and the sampling task are computationally easier as they are performed over a lower dimensional space.

        One of the most challenging aspects is the identification of the, unknown, effective dimensionality. In general the set of linearly independent vectors is approximated by an embedding that is learned by the algorithm~\cite{dalalyan2008new,yin2015sequential,del2020effective}. 
        As a result, projection methods typically differ in: \begin{inparaenum}\item[(1)] the way to identify the, lower dimensional, embedding; 
        \item[(2)] the technique used to implement the projection from the original space onto the embedding; and \item[(3)] the sampling function used.\end{inparaenum} 
        
        
       An example of a projection based approach is the Random EMedding Bayesian Optimization (REMBO) algorithm~\cite{wang2013bayesian,wang2016bayesian,chen2012joint}. In REMBO locations are sequentially sampled according to a distribution defined over the low dimensional embedded space. The algorithm generates a random projection matrix that maps samples from the embedding to the original space and vice versa. Sampling distributions are implied by the Gaussian process surrogate model constructed over the embedded space, and the associated acquisition function. The embedding dimension is specified \textit{a priori}, i.e., no mechanism is provided to learn the embedding dimension. Recent work has investigated approaches to automate the selection of the embedding dimension \cite{binois2020choice}.    
        
        A challenge for projection based approaches is maintaining consistency in the distance among locations when going from the original to the low dimensional subspace. \cite{binois2015warped} proposes a warped Gaussian process kernel to improve on this lock of consistency of the distance metric. HeSBO (Hashing-enhanced sub-space Bayesian Optimization~\cite{nayebi2019framework}) uses two hash functions to implicitly represent the embedding, enabling the constructed mapping to have low distortion, i.e., no point mapped from the embedded domain to the original domain is outside of the original domain. The sub-space identification Bayesian optimization (SI-BO) algorithm utilizes a single projection matrix to create an embedding~\cite{djolonga2013gaussian}, and leverages low-rank matrix recovery techniques to construct the embedding mapping. 
        Once the embedding is generated, Bayesian optimization is used to search over that embedded space, and the Gaussian process Upper Confidence Bound (GP-UCB) is used as the acquisition function.
        
        Approaches have been propose that sequentially update the embedding. As an example,~\cite{qian2016derivative} proposes to incorporate sequential embeddings, rather than a single random embedding, by iteratively generating and optimizing over the embeddings. 
         

        Embedding approaches have gained traction in applications where most dimensions do not significantly impact the objective function~\cite{bergstra2012random}. On the other hand, the assumption of low effective dimensionality has been noted as restrictive in 
        several applications~\cite{kandasamy2015high}. Examples of problem classes where a large number of decision variables affect the objective function include: additive functions, network functions, and non-separable functions. In fact, practical problems where low effective dimensionality does not hold are found across many applications~\cite{murphy1994uci,viola2001rapid,kandasamy2015high,snelson2007local}. 
    
    \subsection{High Dimensional Statistics Based Approaches}\label{sec::sotahdsl}
        
        Statistical modeling approaches attempt to exploit structural properties of the function to be learned. Exploiting such properties can improve the computational tractability of estimating high dimensional surrogate model parameters.
        We first review methods for efficient learning of high dimensional Gaussian processes. These models are relevant to our research, but were not developed within an optimization context. We then look into models developed to optimize a black box function. 

        In the statistical learning community, the focus is in investigating efficient methods for high dimensional modeling~\cite{bozdogan2016novel,pamukcu2019choosing,bettonvil1997searching,lu2020faster}. 
        The approach in~\cite{bettonvil1997searching} relies on the structural properties of linear models, leveraging linear regression equations to solve a simulation input screening problem. 
        \cite{pamukcu2019choosing} uses an adaptive elastic net regularization and a linear model structure. The authors use computationally efficient hybrid covariance estimators and an information complexity criteria to quickly screen through thousands of potential inputs. 
        Recently,~\cite{lu2020faster} proposed to scale Gaussian process models to high dimensions by leveraging work in the related field of Gaussian process estimation over large data sets~\cite{hensman2013gaussian,binois2018practical,kleijnen2020prediction}. The authors build Gaussian process kernels based upon a suitably selected subset of samples and employ Nystr\"{o}m regularization~\cite{rudi2015less} to select a subset of columns to be used for model estimation to efficiently compute Gaussian processes for high dimensional spaces. 
        
        As previously mentioned, these surrogates, were not designed for optimization.
        In this context, the main model form that has been investigated is the additive Gaussian process. Additive Gaussian processes assume that the original, high dimensional, function is the sum of a number of components, where each component only depends on a subset of the problem dimensions~\cite{gyorfi2006distribution}. The use of an additive model structure dramatically improves the tractability of the surrogate likelihood function, which must be maximized to derive the model hyperparameters.
        There have been several proposed methods that use additive Gaussian processes for high dimensional function optimization with differences in: \begin{inparaenum}
        \item[(1)] how the components are learned, \item[(2)] how the additive Gaussian process hyperparameters are estimated, and \item[(3)] how the additive Gaussian process is used to take samples.
        \end{inparaenum}
        
        In their work, \cite{kandasamy2015high} proposes to use additive Gaussian processes for high dimensional Bayesian optimization, and requires the number of model components to be provided as input by the user. 
        Alternative decompositions of the original function into the model components are randomly generated by the algorithm, and the decomposition with the largest associated marginal likelihood is selected.
        The authors consider no overlap among the model components, i.e., no dimension appears in more than one model component. 
        Sampling locations are sequentially generated by extending the Upper Confidence Bound (UCB) criteria~\cite{auer2002using} to the case of Gaussian processes with additive structure. The authors show how, under the proposed additive model structure, the UCB acquisition function is separable and can be optimized over each of the model components. 
        \cite{li2016high} extends the approach of \cite{kandasamy2015high} by using a projection pursuit additive Gaussian process model to alter the marginal likelihood when deciding the model component decomposition. In~\cite{mutny2018efficient}, the authors introduce a Quadrature Fourier Feature (QFF) technique to reduce the time for kernel inversion. 
        
        Generalized additive structures have been proposed and analyzed to consider overlapping model components. 
        \cite{gardner2017discovering} makes use of a MCMC approach to derive the overlapping component decomposition by combining the marginal likelihood calculation with a Metropolis-Hastings step; the expected improvement acquisition function~\cite{jones1998efficient} is used to determine sampling. \cite{rolland2018high} presents a Gibbs sampling algorithm to determine the component decomposition, learning the additive structure over a graph, and UCB is used for sampling.  
        \cite{wang2017batched} makes use of Gibbs sampling to learn the model components. The authors extend additive Gaussian process model based optimization to parallel resources, proposing a batched approach where a UCB acquisition function is paired with a Determinantal Point Process \cite{kathuria2016batched} to select batches of sample points at each iteration. \cite{wang2018batched} builds upon the Gibbs sampling method from the previous work in~\cite{wang2017batched} to select model components, and an ensemble of Mondorian trees is introduced to create an Ensemble Bayesian Optimization.

    \subsection{Contributions}\label{sec::contributions}
        This paper proposes a new approach for Bayesian optimization in high dimensions that does not fall into any of the categories presented in Section~\ref{sec::SoTA}. Specifically, 
        while parallels can be drawn between our approach and embedding or additive modeling, 
        our BOFiP algorithm does not rely on the assumptions these require to be effective.
        BOFiP distributes the optimization creating several sub-problems, each defined in a different feasible sub-space such that the intersection of the sub-problem feasible spaces is the empty set and their union is the original feasible space. 
        The sub-problems in each sub-space are approximately solved using Bayesian optimization. The solutions and associated rewards generate the sub-problems for the next BOFiP iteration. Bayesian optimization is then started again. The optimization/update cycles continue until a stopping criteria is met. 
        
        BOFiP shares aspects with other approaches in the literature, but maintains distinctive features: \begin{inparaenum}\item[(i)] 
        though the optimization over sub set of dimensions is similar to coordinate descent/ascent algorithms, 
        our method differs since the dimensions are fixed to values randomly sampled according to a distribution that is iteratively updated. \item[(ii)] BOFiP extends the game theoretic work in sampled fictitious play~\cite{lambert2005fictitious} to Bayesian optimization. 
        \item[(iii)] 
        Our approach can be paired with ideally any of the methods presented in Section~\ref{sec::sotahdsl}
        \end{inparaenum}. 
        
        A preliminary version of this work was introduced in~\cite{mathesen2019subspace}, where the concept of space decomposition was first presented with preliminary results. Here, we propose an entirely new synchronization mechanism among sub-spaces reducing the input parameters required by the approach, and provide an extended numerical analysis. 

\section{Bayesian Optimization with Sampled Fictitious Play (BOFiP)}
\begin{figure}[H]
        \centering
        {\includegraphics[width=0.45\textwidth]{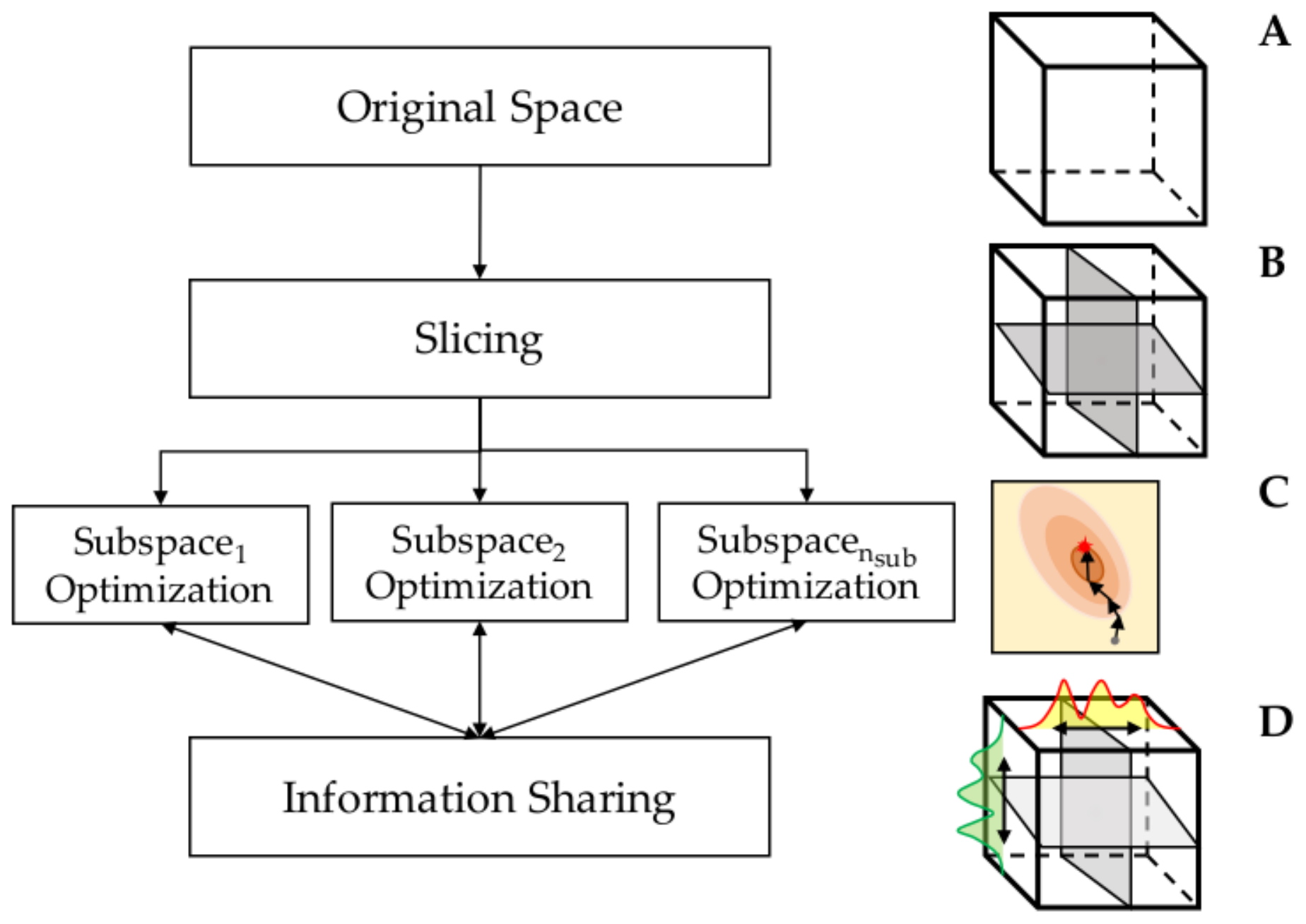}
        \caption{The original space (A) is sliced into \textit{p} sub-spaces of lower dimension (B). Bayesian Optimization is performed, simultaneously, in each sub-space (C). The shared beliefs, that are updated each time a Bayesian optimization is terminated, help in biasing the random sliding of the slices (D).}}
        \label{fig::frameworkBOFiP}
    \end{figure}

    We aim to solve the following optimization problem:
    \begin{eqnarray}\label{eq: problem}
        \mathbf{x}^{*} = \arg \underset{\mathbf{x}\in\mathbb{X}\subset\mathbb{R}^d}{\min}f(\mathbf{x})\label{eqn::prob}
    \end{eqnarray}
    where the solution space $\mathbb{X}$ is high dimensional, i.e., $d\ge 20$, and $f(\mathbf{x})$ is a black box function. We tackle the high dimensional problem by iteratively optimizing over partitions of the dimensions of the original space. Similar to additive modeling, we assign the dimensions to non-overlapping sets (thus creating the sub-spaces, as shown in step B of Figure~\ref{fig::frameworkBOFiP}). Subsequently, similar to the embedding approaches, we execute Bayesian optimization (step C of Figure~\ref{fig::frameworkBOFiP}) in each individual sub-space.
    Once the sub-space optimizations are terminated, BOFiP initiates information sharing among sub-spaces (step D of Figure~\ref{fig::frameworkBOFiP}). Information sharing updates the \textit{sub-space} distribution, which characterizes the probability that a sub-space location is a minimum/maximum of the associated sub-problem. 
    The product of the sub-space distributions is the BOFiP distribution defined over $\mathbb{X}$, which characterizes the probability that a location $\mathbf{x}\in\mathbb{X}$ improves over the current solution. 
    Sub-space objective functions are updated as a result of information sharing; with the updated objectives, the sub-space Bayesian optimization is executed again. 
    
    BOFiP relies on two components: \begin{inparaenum}\item[(i)] a Bayesian optimizer, and 
    \item[(ii)] sampled fictitious play~\cite{lambert2005fictitious} to update the distribution of the optimum location used in sub-space information sharing and objective function update. \end{inparaenum}
    Sections~\ref{sec::GPEI}-\ref{sec::FictPlay} introduce the two components separately, while Section~\ref{sec::BOFiP Algorithm} details the BOFiP algorithm. 

    \subsection{Bayesian optimization with Gaussian processes and expected improvement} \label{sec::GPEI}
        Bayesian optimization conveniently interprets the objective to be minimized as the realization from a random function. In particular, we consider stationary Gaussian processes as surrogates for the, unknown, objective function. At any un-sampled location $\mathbf{x}$, the conditional density of the unknown objective function is normal with parameters dependent on locations and function values at previously sampled points~\cite{huang2006global,santner2013design}. A Gaussian process is defined as $F(\mathbf{x}) = \mu + Z(\mathbf{x})$, where $\mu$ is the, constant, process mean, and $Z(\mathbf{x})\sim GP(0,\tau^2R)$, with $\tau^2$ being the constant process variance and $R$ the correlation matrix. Under the Gaussian correlation assumption, $R_{ij} = \prod_{l=1}^d \exp\left(-\theta_l \left(x_{il}-x_{jl}\right)^2\right)$, for $i, j, = 1,\ldots,n$. The $d$-dimensional vector of hyperparameters $\boldsymbol{\theta}$ controls the 
        smoothing intensity of the predictor in the different dimensions. The parameters $\mu$ and $\tau^2$ are estimated through maximum likelihood~\cite{santner2013design}:
        $\hat{\mu} = \frac{\boldsymbol{1}_n^T\mathrm{R}^{-1}f(\boldsymbol{X}_n)}{\boldsymbol{1}^T_n\mathrm{R}^{-1}\boldsymbol{1}_n}, \ 
        \hat{\tau}^2= \frac{(f(\mathbf{X}_n)-\boldsymbol{1}_n\hat{\mu}_g)^T \mathrm{R}^{-1}((f(\mathbf{X}_n)-\boldsymbol{1}_n\hat{\mu}_g)}{n}$. The best linear unbiased predictor 
        form is~\cite{santner2013design}:
        \begin{eqnarray}
            \hat{f}(\mathbf{x}) = \hat{\mu} + \mathbf{r}^T\mathrm{R}^{-1}(f(\mathbf{X}_n) - \mathbf{1}_n\hat{\mu})\label{eqn::yhat}
        \end{eqnarray}
        where $\mathbf{X}_n$ is a set of $n$ sampled locations, and $f(\mathbf{X}_n)$ is the $n$-dimensional vector having as elements the function value at the sampled locations. The model variance associated to the predictor is:
        \begin{eqnarray}
        s^2(\mathbf{x}) = \tau^2\left(1-\mathbf{r}^T\mathrm{R}^{-1}\mathbf{r} + \frac{(1-\mathbf{1}_n^T\mathrm{R}^{-1}\mathbf{r})^2}{\mathbf{1}_n^T\mathrm{R}^{-1}\mathbf{1}_n} \right)\label{eqn::mvar}
        \end{eqnarray}
        where $\mathbf{r}$ is the $n$-dimensional vector having as elements the Gaussian correlation between location $\mathbf{x}\in\mathbb{X}$ and the $n$ elements of $\mathbf{X}_n$, i.e., $\mathbf{r}_i(\mathbf{x}) =\prod_{l=1}^d \exp\left(-\theta_l(x_l-x_{il})^2\right), i = 1,\ldots,n$. 
        The predictor in~\eqref{eqn::yhat} and the associated model variance in~\eqref{eqn::mvar} are used to derive the \emph{acquisition function} whose maximization determines the next sample point. In this work, we consider the \textit{expected improvement} (EI)~\cite{jones1998efficient} acquisition function. Letting $\mathbf{x}_{*}\in \arg \min_{\mathbf{x}\in \mathbf{X}_n} f(\mathbf{x})$ be the best observed location, and $\Delta_f(\mathbf{x}) = f(\mathbf{x}_*)-\hat{f}(\mathbf{x})$, the EI for any location $\mathbf{x}\in\mathbb{X}$ is: 
        \begin{eqnarray}\label{eqn::EI}
            EI(\mathbf{x}) = \max \bigg\lbrace  \Delta_f(\mathbf{x})\Phi\left(\frac{\Delta_f(\mathbf{x})}{s(\mathbf{x})}\right) + s(\mathbf{x}) \phi\left(\frac{\Delta_f(\mathbf{x})}{s(\mathbf{x})}\right),0 \bigg\rbrace \label{eqn::EIdef}
        \end{eqnarray}
        where the expected improvement is $0$ for any sampled point, i.e., $EI\left(\mathbf{x}\right)=0\; \ \forall \mathbf{x}\in\mathbf{X}_{n}$.
        The next location to be sampled is then determined 
        such that:
        \begin{eqnarray}
        \mathbf{x}_{n+1} \in \arg\max_{\mathbf{x}\in \mathbb{X}}EI(\mathbf{x}).\label{eqn::samplingrecursion}
        \end{eqnarray}
        The algorithm progresses until a stopping condition is met.
        
    \subsection{Sampled Fictitious Play} 
        \label{sec::FictPlay}
        Fictitious play was first introduced in the game theory literature in~\cite{brown1951iterative} and~\cite{robinson1951iterative}. The fictitious play algorithm is a method for the identification of Nash equilibria for games with identical payoffs. These are referred to as \textit{Equal Interest Games}, and represent a special class of \textit{Potential Games}. Following the formalism presented in~\cite{monderer1996fictitious}, let $\Gamma$ be a finite game formulated in strategic form, where the set of players $P=\left\lbrace 1,2,\ldots,p\right\rbrace$ is defined such that each player $i\in P$ has an associated finite set of strategies that she can play, referred to as $\mathcal{Y}^{i}$. The set of feasible plays can then be expressed as $\mathcal{Y}=\mathcal{Y}^{1}\times\mathcal{Y}^{2}\times\ldots\times\mathcal{Y}^{p}$. Then, the \textit{payoff} (utility, function, reward) for the generic player $i\in P$ is referred to as $u^{i}:\mathcal{Y}\rightarrow\mathbb{R}$, and it holds that $u^{1}=u^{2}=\ldots=u^{p}=u$. In the context of game theory, we associate the set of mixed strategies to each player $i\in P$, namely:
        \begin{eqnarray}
        \vspace{-12pt}
        \Delta^{i}=\left\lbrace f^{i}:\mathcal{Y}^{i}\rightarrow\left[0,1\right]:\sum_{y^{i}\in\mathcal{Y}^{i}}f^{i}\left(y^{i}\right)=1\right\rbrace\label{eqn::defBelief}
        \vspace{-12pt}
        \end{eqnarray}
        In definition~\eqref{eqn::defBelief}, 
        we can interpret each $f^{i}\in\Delta^{i}$ as an assignment of probabilities, or \textit{beliefs}, to the elements of the vector of the feasible plays for player $i$, $\mathcal{Y}^{i}$. Several algorithms have been proposed to solve equal interest games and the analysis of fictitious play was discussed in~\cite{young2004strategic}, where a version with player \textit{inertia} was proposed and discussed. More recently,~\cite{swenson2018best} discusses properties for equal interest games within the more general context of potential games. Another two contributions, from the optimization domain, that are strongly related to our work, are~\cite{lambert2005fictitious}, which investigates the properties of a \textit{sampled} version of the fictitious play algorithm, and~\cite{dolinskaya2016parameter}, where a single sample version of the work in~\cite{lambert2005fictitious} is proposed to solve a large scale dynamic programming problem. 
        
        \begin{algorithm}[H]
            \caption{Sampled Fictitious Play: SFP~\cite{lambert2005fictitious}}
            \label{alg::SFP}
            \begin{algorithmic}
                \STATE {\bfseries Initialize:} Set $t=1$ and select $y(1)\in\mathcal{Y}=\mathcal{Y}^{1}\times\mathcal{Y}^{2}\times\ldots\times\mathcal{Y}^{p}$ arbitrarily; set $f_y(1)=y(1)$.
                \STATE \textbf{Step 1}: Given $f_y(t)$, select a sample size $k_t\ge1$, and draw i.i.d. random samples $Y_j(t), \ j=1,\ldots,k_t$, from the distribution given by $f_y(t)$. 
                \STATE \textbf{Step 2}: Using the above sample, find
                \begin{align}\label{eq:SFP opt}
                    y^i(t+1) \in \underset{y^i\in \mathcal{Y}^i}{\arg \max}\{ \bar{u}^i_{k_t}(y^i,f_y^{-i}(t))\}, \quad i=1,\ldots,p.
                \end{align}
                Where $\bar{u}^i_{k_t}(y^i,f_y^{-i}(t))$ is a realization of the sample average $$\bar{U}^i_{k_t}(y^i,f_y^{-i}(t)) = \sum_{j=1}^{k_t}\frac{u^i(y^i,Y^{-i}_j(t))}{k_t}$$
                and $Y^{-i}_j(t),\ j = 1,\ldots,k_t$, are i.i.d. random vectors with distribution $f_y^{-i}(t)$. 
                \STATE \textbf{Step 3}: Set
                \begin{align}\label{eq:SFP update}
                    f_y(t+1) = f_y(t)+\frac{1}{t+1}\left(y(t+1)-f_y(t)\right)
                \end{align}
                \STATE \textbf{Step 4}: $t\leftarrow t+1$, return to Step 1.
           \end{algorithmic}
        \end{algorithm}
        
        However, all of the previous approaches to fictitious play assume that each player has access to the utility function form and can calculate the exact best reply to the actions of all the other players. Specifically, in~\cite{monderer1996fictitious}, at each iteration, the best reply $y^i$ is updated as $y^i(t+1) \in \underset{y^i\in \mathcal{Y}^i}{\arg \max}\ u^i(y^i,f_y^{-i}(t))$, where $f_y^{-i}$ represents the probability assignment beliefs of all players other than $i$, such that the optimum play (solution) is discovered and utilized at each iteration. \cite{lambert2005fictitious} extends~\cite{monderer1996fictitious} to the sampled belief case by considering only a subset of possible plays and calculating the exact best reply to the empirical reward, i.e., $\underset{y^i\in \mathcal{Y}^i}{\max}\ E_{Y^{-i}}(u^i(y^i,Y^-i))$, where $Y^{-i}$ is a random vector with components $Y^j$, $j\neq i$, that have probability distribution $f_y^{j}$.~\cite{lambert2005fictitious} shows that the updates to $y^i$ can be made over a sample average as described in Algorithm~\ref{alg::SFP}.

        This update of the sample average of the belief vector requires the ability to calculate the \textit{exact} best reply to the sample realizations $\bar{u}^i_{k_t}(y^i,f_y^{-i}(t))$, as seen in~\eqref{eq:SFP opt}.
        With respect to our black box optimization problem, the problem in~\eqref{eq:SFP opt} cannot be solved exactly. 
        In BOFiP, we use the sampled fictitious play belief vector update step in~\eqref{eq:SFP update}, but we circumvent the impossibility to calculate the best reply to the empirical utility~\eqref{eq:SFP opt} by computing an approximate solution using a Bayesian optimizer.
        
    \subsection{BOFiP Algorithm}\label{sec::BOFiP Algorithm} 
    
        The idea at the basis of BOFiP to scale black box optimization to high dimensional problems is to interpret the minimization in~\eqref{eq: problem} as an equal interest game, which we solve using an approximation of the sampled fictitious play algorithm (Section~\ref{sec::FictPlay}). 
        To search for the minimum of the acquisition function in~\eqref{eqn::EIdef} (which is highly nonlinear and non-convex), we consider a fine grid of the original space $\mathbb{X}$. In the context of BOFiP, the domain discretization makes the application of fictitious play natural: each  location in the original space can be interpreted as a combination of strategies from the individual players (sub-spaces). 
        We acknowledge that implementations of fictitious play in continuous action spaces have also been investigated~\cite{perkins2014stochastic}, and could be used to extend BOFiP.
        
        BOFiP decomposes the original high dimensional domain $\mathbb{X}$ into $p$ sub-spaces, $\mathbb{P} = \{1,\ldots, p\}$. 
        Each sub-space has dimensionality $d^{i}<< d$ (the sub-problem complexity is controlled by choosing $d^i$). Let us refer to the set of dimension indexes that each sub-space contains as $I^{i}, \ i=1,\ldots,p$. The sub-space decomposition is such that $\cap^{p}_{i=1}I^{i}=\emptyset$, and, letting $D=\left\lbrace 1,2,\ldots, d\right\rbrace$ be the set of all dimension indexes, we have $\cup^{p}_{i=1}I^{i}\equiv D$. 
        We will refer to each $d^{i}$-dimensional sub-space as $\mathbb{X}^i$, and we will refer to the feasible space, excluding the $i$\textsuperscript{th} sub-space, as $\mathbb{X}^{-i}$. By construction, we have that $\mathbb{X} = \mathbb{X}^1 \times\ldots\times\mathbb{X}^p$ resembles the game setup presented in Section~\ref{sec::FictPlay}. 
        Each sub-space is fully characterized by the discrete set of feasible locations $\mathbf{x}^{i}_{j}\in\mathbb{X}^{i}$, where $\mathbf{x}^{i}_{j}$ is a $d^{i}$-dimensional vector and we have a number of locations, given by the discretization, $j=1,2,\ldots,|\mathbb{X}^{i}|$.
        
        Each sub-space $\mathbb{X}^i$ maintains a $d^{i}$-dimensional belief vector $f^{i}_{b}$ 
        over its set of feasible locations $\mathbb{X}^i$ that, following Section~\ref{sec::FictPlay}, captures the likelihood that a specific location is also an equilibrium for the original game. 
        The belief vectors $f^{i}_{b}, i=1,2,\ldots,p$, are shared and used by each sub-space $j$, $j\neq i$, as part of the learning strategy to ``simulate'' the behavior of other sub-spaces. 
        Specifically, each time a location $\mathbf{x}\in \mathbb{X}^i$ is sampled within the $i$\textsuperscript{th} sub-space, to evaluate the black box function $f(\mathbf{x})$, a value for $\mathbf{x}^{-i}\in\mathbb{X}^{-i}$ is needed. In order to do so, the $i$\textsuperscript{th} player samples from the joint belief density of the remaining players, i.e., $f^{-i}_{b}\left(\mathbf{x}^{-i}\right) = \prod_{j\neq i}f^{j}_{b}\left(\mathbf{x}^{j}\right)$. In fact, the implied belief vector over the original domain for each location belonging to the grid is $f_{b}\left(\mathbf{x}\right) = \prod^{p}_{i=1}f^{i}_{b}\left(\mathbf{x}^i\right)$. 
        
        Note that our sub-spaces remarkably differ from an embedding. In the embedding literature, high dimensional points are mapped into a single lower dimensional space. In our approach, the $p$ low dimensional sub-spaces cover the high dimensional space, and ``communicate'' via the belief vectors $f^{i}_{b}, i=1,2,\ldots,p$.  
        
        Following the initial decomposition of the original domain into sub-spaces, the belief vector associated to each of the sub-spaces is initialized to be a discrete uniform distribution over the respective sub-space, i.e., with a slight abuse of notation, $f^i_{b}(t=0)\equiv \mbox{DU}\left(\mathbb{X}^{i}\right)$, $t$ being the iteration index (
        this initialization is the first three steps of Algorithm~\ref{alg:: SCOOP}). 
        At the $t$\textsuperscript{th} BOFiP iteration, the ``local'' optimizations in each sub-space $i\in \mathbb{P}$ requires to draw $k_t$ i.i.d. locations $\mathbf{x}^{-i}(t)\in\mathbb{X}^{-i}$ (Step 1 of Algorithm~\ref{alg::SFP}). 
        The $k_t$ samples from each of the $p$ belief vectors at the $t$\textsuperscript{th} iteration are referred to as $\mathbf{x}^{-i}_h(t)$ for $h = 1,\ldots,k_t$. 
        Randomly sampling $\mathbf{x}^{-i}_h(t)$ from $f^{-i}_b(t)$ 
        is the primary exploration mechanism of sampled fictitious play~\cite{lambert2005fictitious}. 
        Once $\mathbf{x}^{-i}_h(t)$, $h = 1,\ldots, k_t$ have been drawn, for each sub-space $i \in \mathbb{P}$, 
        Bayesian optimization is performed (Step 5 in Algorithm~\ref{alg:: SCOOP}). At each iteration of the Bayesian optimization, a Gaussian process is estimated and the EI acquisition function is evaluated (equation~\eqref{eqn::EI}) over $\mathbb{X}^i$ to sequentially determine the next point to evaluate. Upon termination of the Bayesian optimization step, an approximate solution to the sub-problem is returned. In fact, the Bayesian optimization (Step 5 in Algorithm~\ref{alg:: SCOOP}) replaces the exact best reply calculation in sampled fictitious play (computed in~\eqref{eq:SFP opt}, Step 2 of Algorithm~\ref{alg::SFP}). 
        
\begin{algorithm}
       \caption{Bayesian Optimization with Sampled Fictitious Play: BOFiP}
       \label{alg:: SCOOP}
    \begin{algorithmic}
       \STATE {\bfseries Input:} domain $\mathbb{X}\subset \mathbb{R}^d$, function $f$, number of sub-spaces $p$, Bayesian Optimization budget $b_{t}=B$; SFP budget $k_{t}$; $\mathbb{X}^{i*}_{\mbox{\tiny{BO}}}\leftarrow \emptyset, \forall i$; BOFiP Iteration Budget $T$; 
       \STATE {\bfseries Output:} Best Location $\mathbf{x}^{*}_{\mbox{\tiny{BOFiP}}}\in\mathbb{X}$
       \smallskip
       \STATE \textbf{Step 1}: Create set of sub-spaces $\mathbb{P} = \{1,\ldots,p\}$, each containing a number of dimensions $d^{i}$ by sampling from the index set $D=\left\lbrace 1,2,\ldots, d\right\rbrace$ without replacement.
       \STATE \textbf{Step 2}: The grid of locations for each sub-space $\mathbb{X}^i, \ i\in \mathbb{P}$, satisfies $\mathbb{X}=\mathbb{X}^1\times\ldots\times\mathbb{X}^p$. Set $t \leftarrow 0$.
       \STATE \textbf{Step 3}: For all $i\in \mathbb{P}$ initialize belief vector $f^i_{b}(t)\equiv \mbox{DU}\left(\mathbb{X}^{i}\right)$\;
       \WHILE{$t<T$}
           \FOR{$i=1,\ldots,n$}
           \STATE \textbf{Step 4}: Sample $k_{t}$ locations, $\mathbf{x}^{-i}_{h}(t), \ h=1,\ldots, k_{t}$, using $f^{-i}_{b}$, 
           \STATE \textbf{Step 5}: Execute \textbf{$\mathbf{x}^{i*}(t)=\mbox{BO}\left(b_{t},\mathbb{X}^{i},\mathbb{X}^{i*}_{\mbox{\tiny{BO}}},\left(\mathbf{x}^{-i}_{h}(t)\right)^{k_{t}}_{h=1}\right)$}\; 
           \STATE \textbf{Step 6}: Update $\mathbb{X}^{i*}_{\mbox{\tiny{BO}}}\leftarrow \mathbb{X}^{i*}_{\mbox{\tiny{BO}}}\cup \mathbf{x}^{i*}\left( t\right)$\;
           \STATE \textbf{Step 7}: Execute update  \textbf{$f^{i}_{b}(t+1)\leftarrow\mbox{BU}\left(f^{i}_{b}(t),\mathbf{x}^{i*}\left( t\right)\right)$}\;
           \ENDFOR
           \STATE \textbf{Step 8}: $t \leftarrow t+1$\;
       \ENDWHILE
       \STATE \textbf{Step 9}: Report best observed location $\mathbf{x}^{*}_{\mbox{\tiny{BOFiP}}}$.
    \end{algorithmic}
\end{algorithm}
        
\begin{algorithm}
    \caption{Bayesian Optimization: \textbf{BO}}
    \label{alg::BOSearch}
    \begin{algorithmic}
        \STATE {\bfseries Input:} Function $f$; BO iteration index $q\leftarrow 0$; sub-space index $i$; BO budget $b_{t}$; Solution set $\mathbb{X} = \mathbb{X}^{i}$; initial set of sampled points; $\left(\mathbf{x}^{-i}_{h}(t)\right)^{k_{t}}_{h=1}$; $\mathbb{S}_{0}=\mathbb{X}^{i*}_{\mbox{\tiny{BO}}}$\;
        \STATE {\bfseries Output:} Best location in $b_{t}$ BO iterations, $\mathbf{x}^{i*}\in \mathbb{S}_q$
        \IF{$\mathbb{S}_{0}\neq\emptyset$}
            \STATE Evaluate the function at $\mathbf{x}^i \in \mathbb{S}_{0}$ using $\Tilde{f}^i_t(\mathbf{x}^i) = \frac{1}{k_t}\sum_{h=1}^{k_t}f\left(\mathbf{x}^i,\mathbf{x}^{-i}_h(t)\right)$, 
            \STATE Estimate $\Tilde{f}^i_t\left(\mathbf{x}^{i}\right)$ using GP at $ \mathbf{x}^{i}\in \mathbb{S}_{0}$, resulting in $\left({\hat{\Tilde{f}}^i_t}_{q}(\mathbf{x}), \hat{s}^{2}_{q}(\mathbf{x})\right)$ for all $\mathbf{x}\in\mathbb{X}$ 
            \STATE Update remaining BO budget $b_{t}\leftarrow b_{t}-k_{t}\cdot|\mathbb{S}_{0}|$
        \ELSE
            \STATE populate $\mathbb{S}_{0}$ with $r_{0}$ random locations, Go back to IF\;
        \ENDIF
        \WHILE {$b_{t}>0$}
	        \STATE $\mathbf{x}^{\mbox{\tiny{EI}}}_{q} \leftarrow \arg \max_{\mathbf{x}\in \mathbb{X}/\mathbb{S}_{q}} \mbox{EI}\left({\hat{\Tilde{f}}^i_t}_{q}(\mathbf{x}), \hat{s}^{2}_{q}(\mathbf{x})\right)$
	        \STATE $\mathbb{S}_{q}\leftarrow \mathbb{S}_q \cup \mathbf{x}^{\mbox{\tiny{EI}}}_{q}$ \;
		\STATE Evaluate the function at $\mathbf{x}^{\mbox{\tiny{EI}}}_{q}$ as\\ $\quad
		\Tilde{f}^i_t(\mathbf{x}^{\mbox{\tiny{EI}}}_{q}) = \frac{1}{k_t}\sum_{h=1}^{k_t}f\left(\mathbf{x}^{\mbox{\tiny{EI}}}_{q},\mathbf{x}^{-i}_h(t)\right)$\; 
		\STATE Update remaining BO budget $b_{t}\leftarrow b_{t}-k_{t}$\;
	    \STATE Update GP estimation using $\mathbb{S}_q$; $\left({\hat{\Tilde{f}}^i_t}_{q}(\mathbf{x}), \hat{s}^{2}_{q}(\mathbf{x})\right)$\;
	    \STATE Update BO iteration index, $q \leftarrow q+1$\;
	    \ENDWHILE
    \end{algorithmic}
\end{algorithm}

\begin{algorithm}
    \caption{Belief Update: \textbf{BU}}
    \label{alg::BU}
    \begin{algorithmic}
        \STATE {\bfseries Input:} Iteration index $t$; sub-space index $i$; belief vector $f^i_{b}$, sub-space $\mathbb{X}^i$, BO locations $\mathbf{x}^{i}\left(t\right)$\;
        \STATE {\bfseries Output:} Updated belief for the $i$\textsuperscript{th} sub-space $f^i_{b}$\;
        \STATE $f^{i}_{b}\leftarrow f^{i}_{b} + \frac{1}{t+1}\left(\mathbf{x}^{i}(t)-f^{i}_{b}(t)\right)$
   \end{algorithmic}
\end{algorithm}
        
        More specifically, at the $t$\textsuperscript{th} iteration, for each of the $\mathbb{X}^i$, $i \in \mathbb{P}$, sub-spaces, given $k_t$ samples $\{\mathbf{x}^{-i}_h(t),\ h=1,\ldots,k_t\}$, a sub-space objective function $\Tilde{f}^i_t(\cdot)$ is defined. The deterministic sub-space objective function optimized by Bayesian optimization, for the $i$\textsuperscript{th} sub-space at the $t$\textsuperscript{th} iteration, is defined as: 
        \begin{align}\label{eq:sub-problem obj func}
            \Tilde{f}^i_t(\mathbf{x}^i) = \frac{1}{k_t}\sum_{h=1}^{k_t}f\left(\mathbf{x}^i,\mathbf{x}^{-i}_h(t)\right).
        \end{align}
        Function~\eqref{eq:sub-problem obj func} is deterministic because $\mathbf{x}^{-i}_h(t),\forall h$ is a function of the BOFiP iteration $t$, but it is kept fixed during the Bayesian optimization. 
        $f\left(\mathbf{x}^i,\mathbf{x}^{-i}_h(t)\right)\equiv f\left(\mathbf{x}\right)$, where $f\left(\mathbf{x}\right)$ is the function we want to minimize in equation~\eqref{eqn::prob}.
        Hence, the deterministic $\Tilde{f}^i_t(\mathbf{x}^i)$ is interpreted as a realization of a Gaussian process. 
        Algorithm~\ref{alg::BOSearch} shows how Bayesian optimization is used to find a solution: 
        \begin{align}\label{eq:local subproblem}
            \mathbf{x}^{i*}(t) \in \arg\underset{\mathbf{x}^i\in\mathbb{X}^i}{\min}\Tilde{f}^i_t(\mathbf{x}^i).
        \end{align}

        Once $b_t\ge0$, iterations of Bayesian optimization are performed, the best solution found for each sub-space, $\mathbf{x}^{i*}(t)$, is stored in $\mathbb{X}^{i*}_{\mbox{\tiny{BO}}}$, i.e., the set of the $i$\textsuperscript{th} sub-space best observed locations. 
        The found sub-space minimum $\mathbf{x}^{i*}(t)$ is used by Algorithm~\ref{alg::BU} to update the respective sub-space belief vector, similar to~\eqref{eq:SFP update}, namely:
        \begin{eqnarray}
        f^{i}_{b}(t+1) = f^{i}_{b}(t) + \frac{1}{t+1}\left(\mathbf{x}^{i*}(t)-f_{b}^i(t)\right).
        \end{eqnarray}
        After the belief update, the next iteration begins ($t=t+1$); newly drawn $\mathbf{x}^{-i}_h(t)$ values are fixed, the associated deterministic sub-space objective functions are updated, and the Bayesian optimization step is re-entered. The initial sampling design for Bayesian optimization over $\mathbb{X}^i$ is augmented by $\mathbb{X}^{i*}_{\mbox{\tiny{BO}}}$ such that the previously learned sub-space solutions are embedded into future sub-space optimizations. 
        
        \begin{figure*}[t]
        	\begin{center}
            	\subfigure[Example GP estimation of $\tilde{f}^{2}_{2}$, $t=2$. \label{fig:exampleGP2}]{\includegraphics[width=0.22\textwidth]{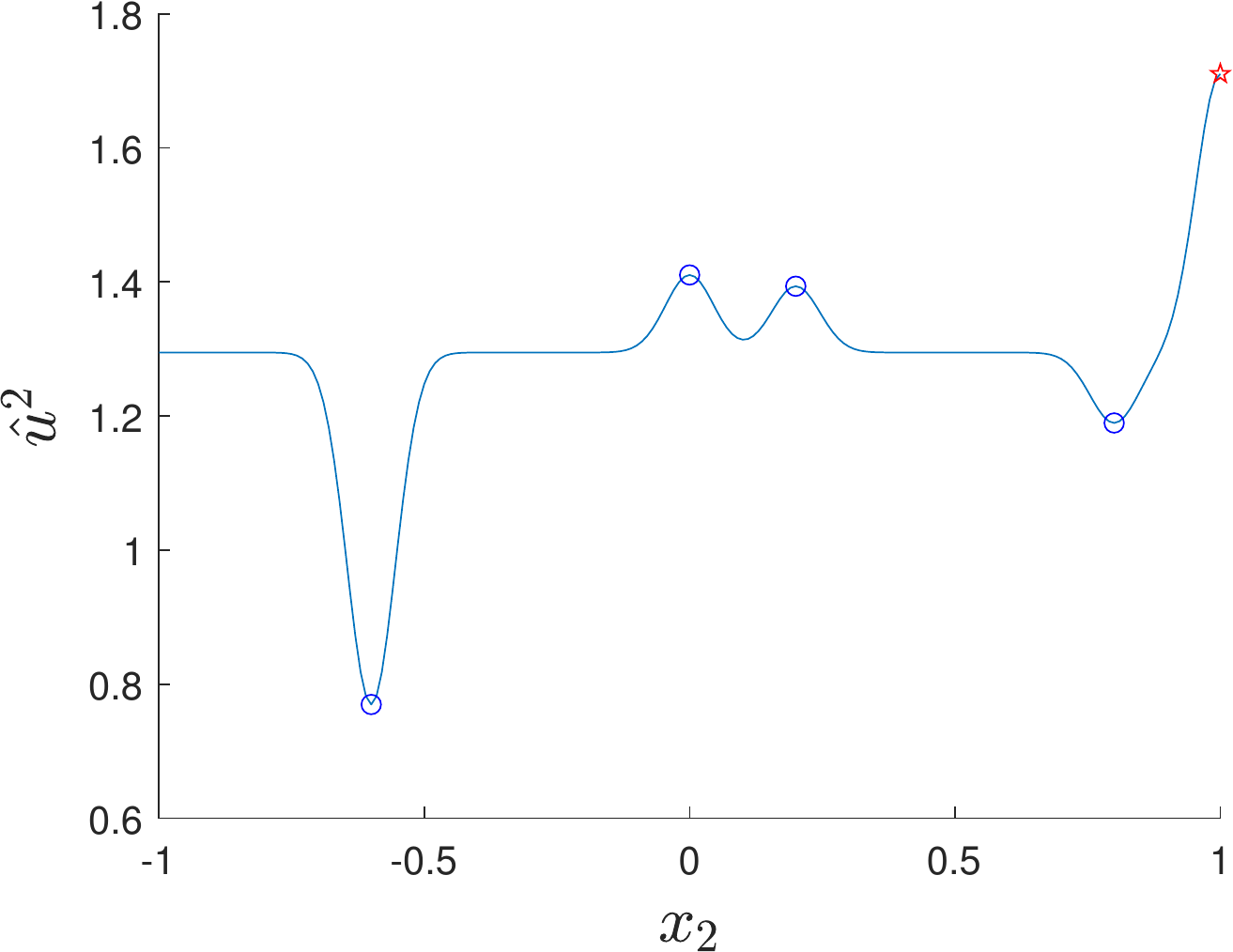}}\ \ 
                \subfigure[Example GP estimation of $\tilde{f}^{2}_{3}$, $t=3$. \label{fig:exampleGP3}]{\includegraphics[width=0.22\textwidth]{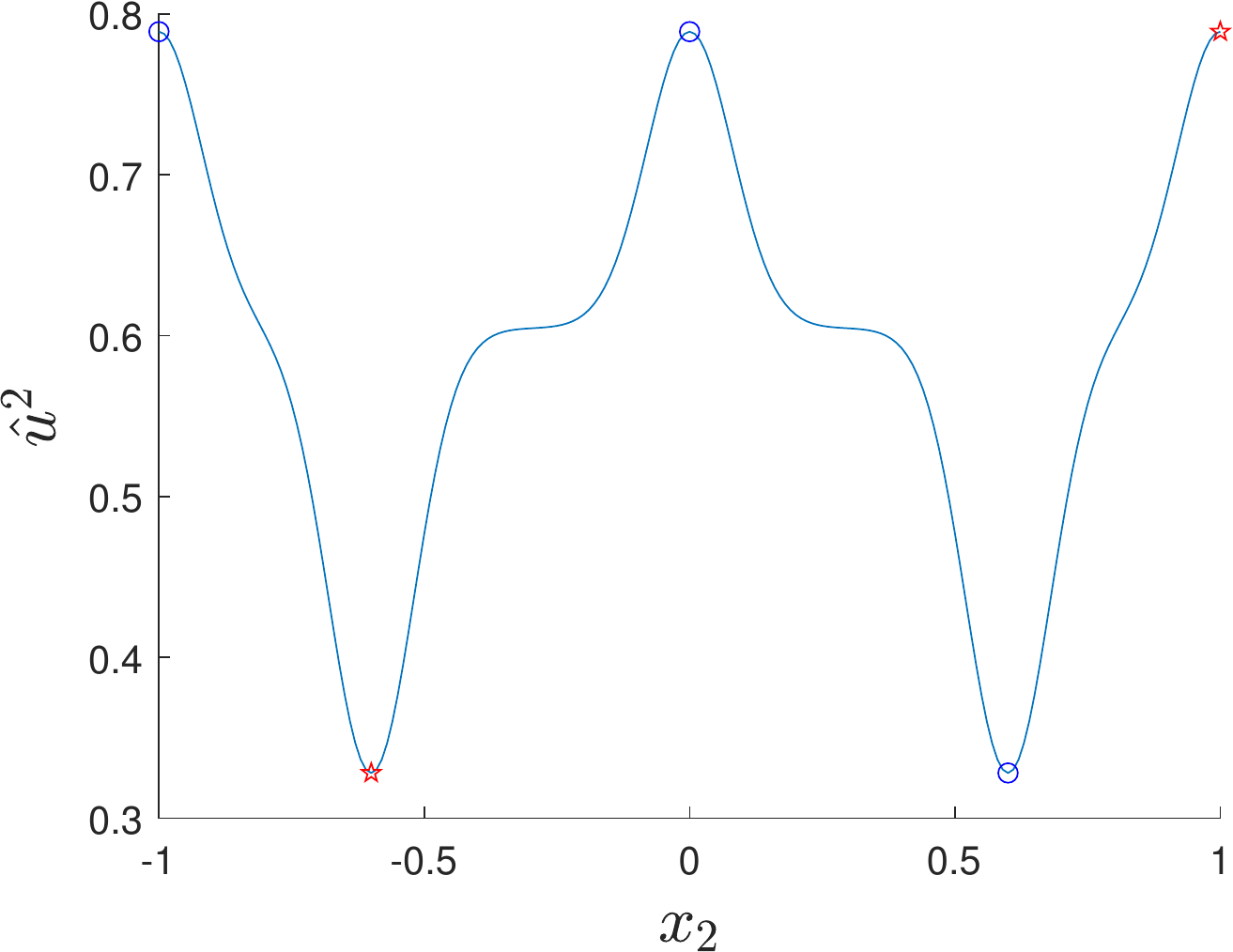}}\ \ 
                \subfigure[Example GP estimation of $\tilde{f}^{2}_{4}$, $t=4$. \label{fig:exampleGP4}]{\includegraphics[width=0.22\textwidth]{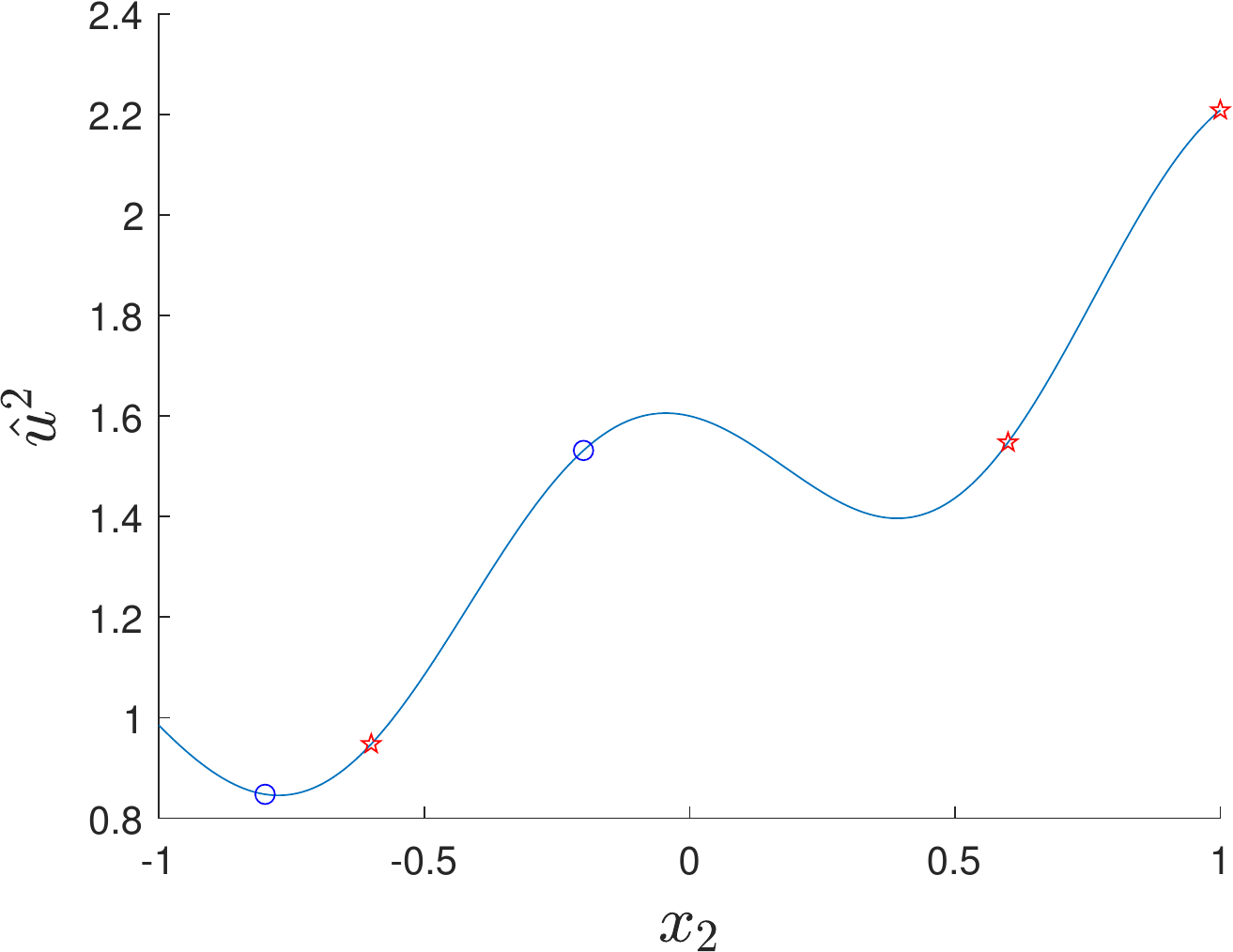}}\ \ 
                \subfigure[Example GP estimation of $\tilde{f}^{2}_{5}$, $t=5$. \label{fig:exampleGP5}]{\includegraphics[width=0.22\textwidth]{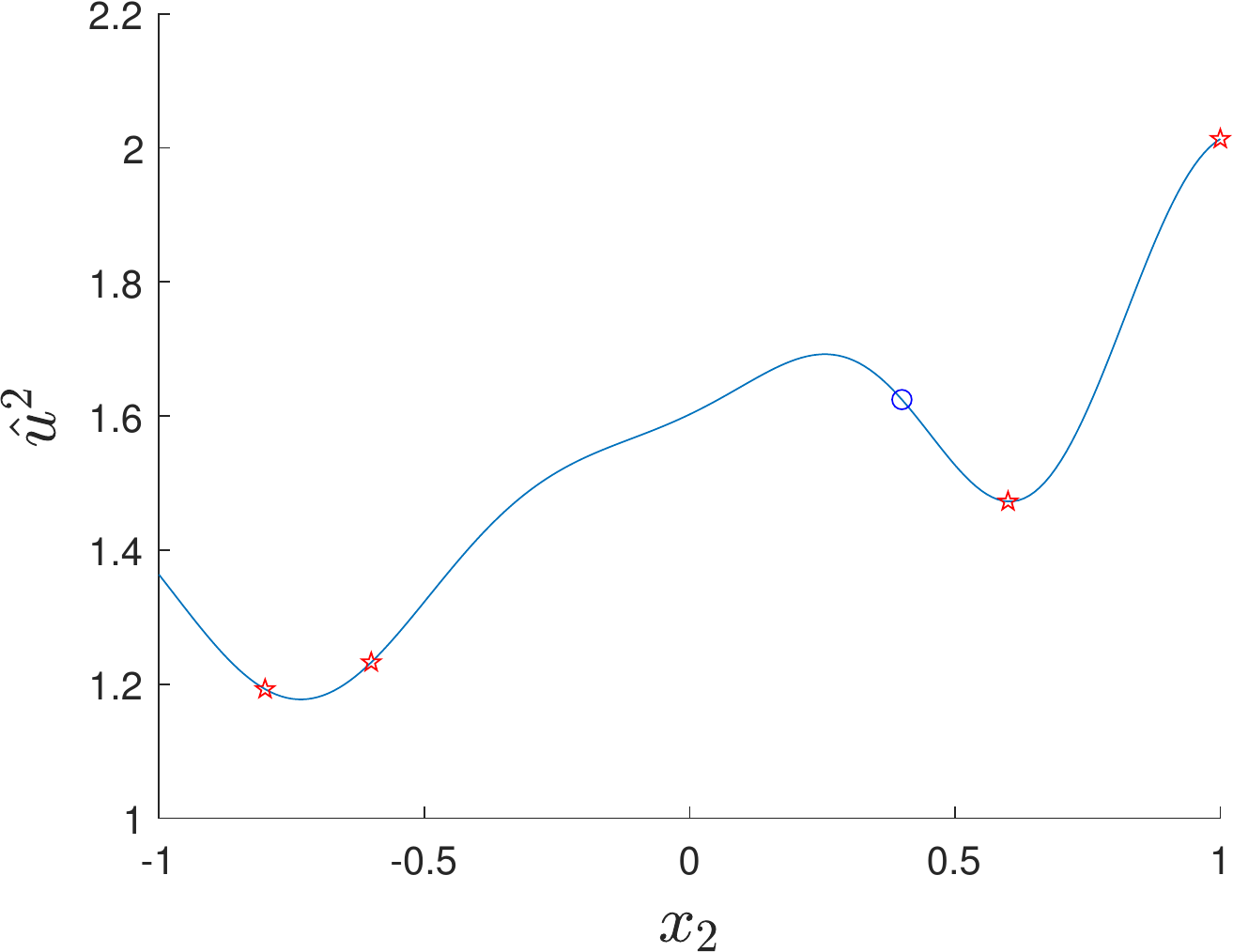}}\ \ 
                \subfigure[Example $f_b^2$, $t=2$. \label{fig:exampleDist2}]{\includegraphics[width=0.22\textwidth]{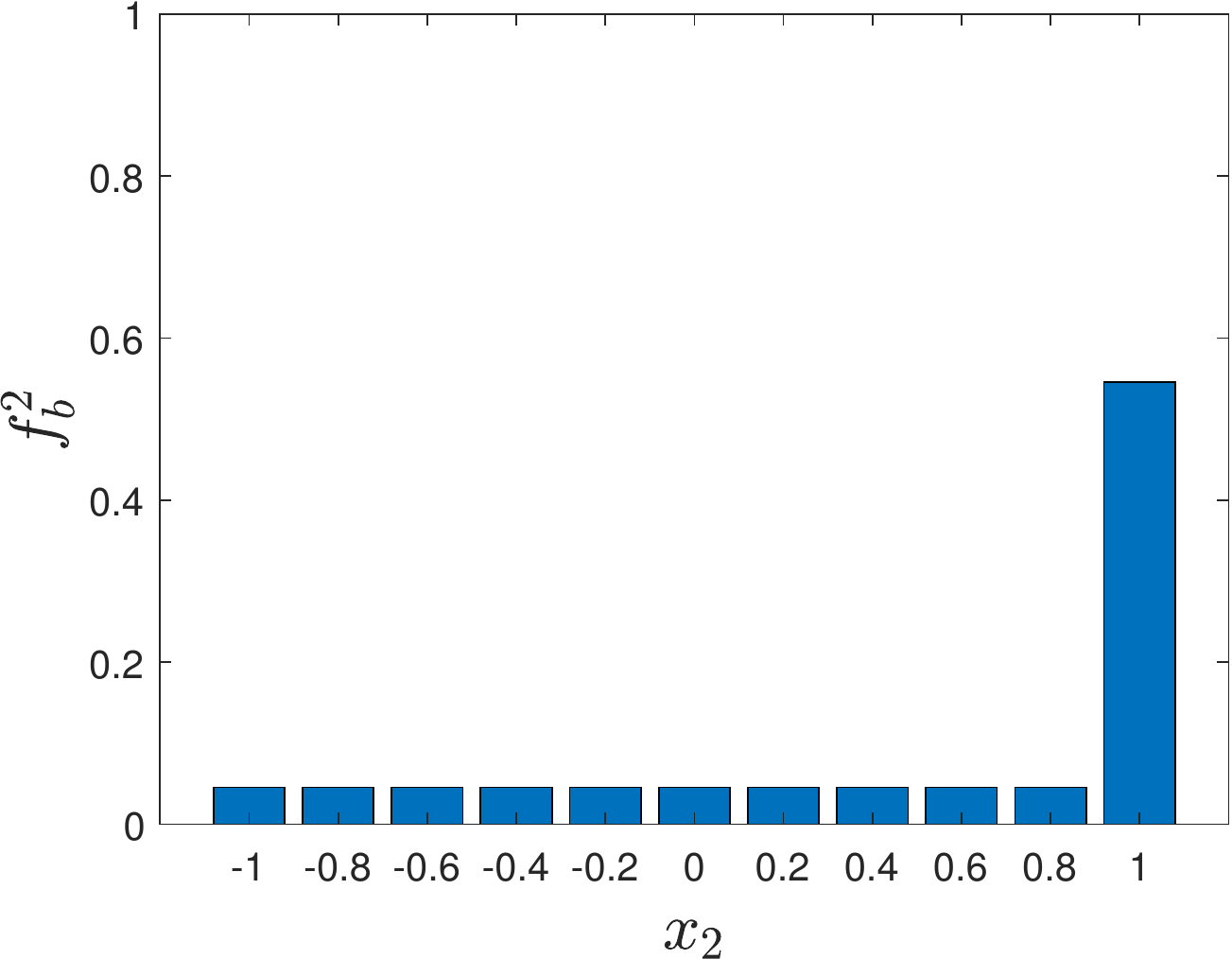}}\ \ 
                \subfigure[Example $f_b^2$, $t=3$. \label{fig:exampleDist3}]{\includegraphics[width=0.22\textwidth]{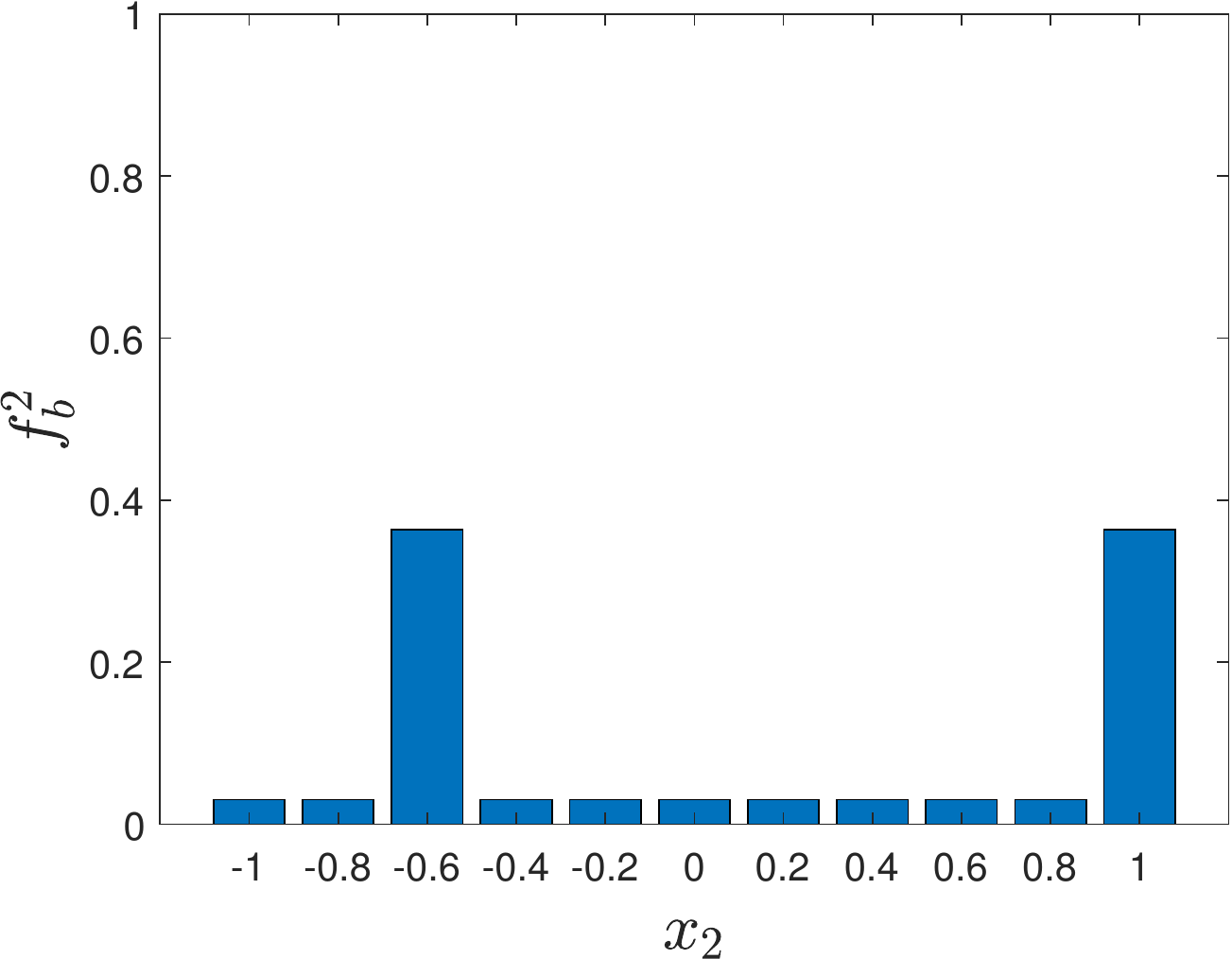}}\ \ 
                \subfigure[Example $f_b^2$, $t=4$. \label{fig:exampleDist4}]{\includegraphics[width=0.22\textwidth]{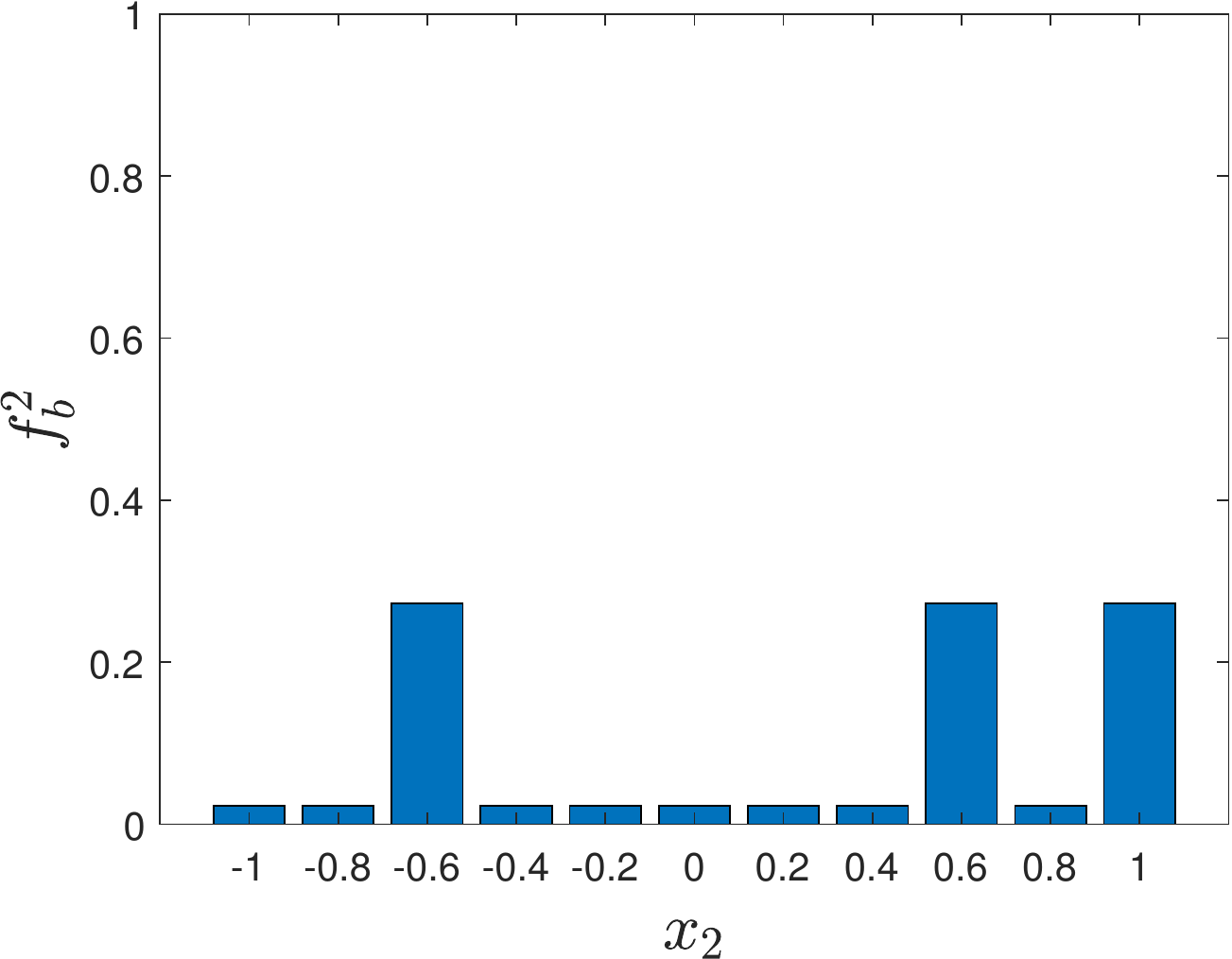}}\ \ 
                \subfigure[Example $f_b^2$, $t=5$. \label{fig:exampleDist5}]{\includegraphics[width=0.22\textwidth]{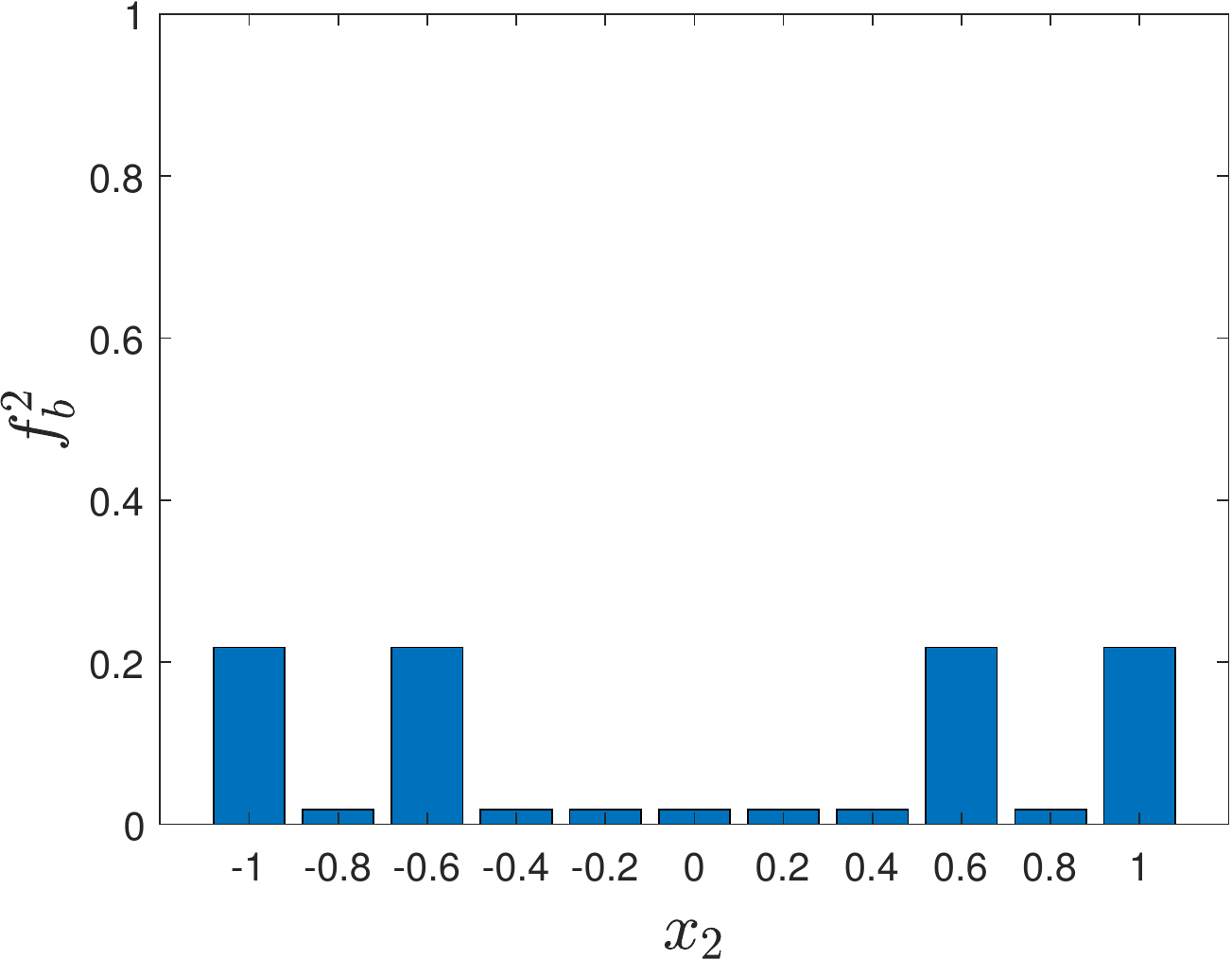}}\ \  
                \caption{Four iterations of BOFiP, red stars are the previous best observations in $\mathbb{X}^{2*}_{\mbox{\tiny{BO}}}$, blue circles are current samples taken during iteration $t$. \label{fig:Example_FP_Evolution}}
        	\end{center}
        \end{figure*}
        
        Figure~\ref{fig:Example_FP_Evolution} shows four iterations of BOFiP, $t=2,\ldots,5$. In this example, $\mathbb{X}=[-1,1]^2$; $p=2$ with one dimension assigned to each sub-space and 
        $\left(\mathbf{x}^{-i}_{h}(t)\right)^{k_{t}}_{h=1}$ sampled from the complementary sub-space's belief vectors.
        Figures~\ref{fig:exampleGP2}-\ref{fig:exampleGP5} show the final Gaussian process predictions from sub-space $i=2$, with red stars indicating samples from $\mathbb{X}^{2*}_{\mbox{\tiny{BO}}}$, and blue circles being the new samples taken at iteration $t$. Figures~\ref{fig:exampleDist2}-\ref{fig:exampleDist5} illustrate the associated belief vector for the same, $i=2$, sub-space. Note that the red stars correspond to the previous best solution found by the Bayesian optimization step, and the belief vector embeds this information. During the second iteration, with only one Bayesian optimization optimum identified, i.e., $\lvert\mathbb{X}^{2*}_{\mbox{\tiny{BO}}}\rvert = 1$, the belief vector has increased density at that location (Figure~\ref{fig:exampleDist2}). However, as more iterations are performed and different candidates are discovered by Bayesian optimization, the belief vector starts to reflect the presence of multiple promising locations. 

        \subsection{Computational Complexity}
        The computational storage space needed to create a dense grid over a high dimensional space is practically infeasible. 
        BOFiP addresses this problem by storing a single grid over each low dimensional sub-space. 
        BOFiP does not explicitly store a full grid in the original $d$ dimensional space, but rather stores a grid of $n^{i}_{g},\, i=1,\ldots,p$ points over each sub-space, resulting in $p$ matrixes of size $n^{i}_{g} \times d^i, i=1,\ldots,p$. Thus a grid in the $d$ dimensional space is implied with $\prod^{p}_{i=1}n^{i}_{g}$ locations. The advantage is that we do not need to store all combinations but only the sub-space generator matrixes, i.e., we only store $p$ matrixes each with $n^{i}_{g}\times d^{i}$ elements. 
         
        
        Bayesian optimization algorithms scale as $\mathcal{O}(m^2)$ (with a $\mathcal{O}(m^3)$ preparatory inversion step), where $m$ is the number of function evaluations~\cite{shahriari2016taking}. This is controlled in BOFiP by means of the Bayesian optimization budget $b_{t},t>0$ at each iteration. 
        Hence, 
        BOFiP complexity is bounded by $\mathcal{O}\left(p\left[\sum_{t}\left(b_t+t-1\right)^2\right]\right)$, where $b_t+t-1$ is the number of samples allocated to the $t$\textsuperscript{th} BOFiP step. 

\section{Experiments}
\label{sec:: Experiments}
\noindent \textbf{Objective of the study.} We performed two sets of experiments to compare BOFiP against state-of-the-art high dimensional Bayesian optimization algorithms to understand the efficiency and applicability of BOFiP. 
In the first set of experiments, we compare the performance when solving non-convex problems, where the location and value of the global minimum are known. 
Results show the behavior of BOFiP as a function of the computational time and difficulty of the problem, expressed in terms of the dimensionality and shape of the function. The second set of experiments 
focuses on the problem of neural network (NN) design, where the number of layers, nodes for each layer, and the weights for each node in the neural network are the decision variables used to optimize and identify the ``optimal'' network configuration. The specific neural network considered was trained to classify breast tumors into two classes, cancerous or benign, based upon tumor feature data~\cite{murphy1994uci}. 
The goal of this part of the experimental study is to understand the behavior of BOFiP when applied to NN training, and to observe how the performance scales as a function of the size of the NN (number of weights).

\noindent\textbf{Metrics.}  We consider the average best observed function value optimality gap across $100$ macro-replications for each experiment, i.e., $\widebar{|f_i-f^*|}$ where $f_i$ is the best solution from the $i$\textsuperscript{th} replication, and $f^{*}$ is the known global optimum function value. This metric represents the average absolute evaluation error that characterizes the algorithm as a function of the wall-clock time. We also report the associated average euclidean distance between the identified and true optimum $\widebar{\lVert\mathbf{x}_i-\mathbf{x}^*\rVert}$. For the case of the neural network, where $f^*$ is unknown, we report $\widebar{|f_i|}$ as the average best observed mean squared error (MSE) over the 100 macro-replications. The best observed MSE for the $i$\textsuperscript{th} macro-replication is calculated as: $f_i = \min \frac{1}{\Omega} \sum_{w=1}^{\Omega}(y_w-\hat{y}_w^i)^2$, where $y_w\in\{0,1\}$ is the true benign/cancerous classification, $\hat{y}_w^i\in\mathbb{R}$ is the predicted classification value for the $w$\textsuperscript{th} data point, and $\Omega$ is the size of the training set.

\subsection{Results}
We compare different high dimensional Bayesian optimization methods to BOFiP on three benchmark functions: the Repeated Branin, and Repeated Hartmann, where each of the $d$ dimensions is constrained in the interval $[-1,1]$, 
and the Ackley function with each dimension constrained within the interval $[-32,32]$~\cite{eggensperger2013towards,oh2018bock,nayebi2019framework}.
In particular, we shift the Ackley function in a way to randomly locate its minimum within the $[-16,16]^{d}$ hypercube. 

We ran the experiments in $20,\ 50,\ 100,$ and $1000$ dimensions. Similar to~\cite{oh2018bock}, we analyze the performance of the algorithms as a function of wall-clock time. 
Following the author's setup, we place a wall-clock time constraint of $0.5$ hours, $1$ hour, $2.5$ hours, and $24$ hours as the stopping condition for execution of the $20$, $50$, $100$, and $1000$ dimension experiments, respectively. To ensure fairness across competitors we ran all the experiments on $4$-core Intel Broadwell CPU machines with 18 GB of RAM. Due to the dramatic increase in problem size for the 1000 dimensional case the RAM allotted to the 4 core Intel Broadwell CPU machines was increased to 40 GB. 
    
\noindent \textbf{Competitors.} We compare three additive Gaussian process-based algorithms~\cite{rolland2018high,gardner2017discovering,wang2017batched} making use of the publicly available implementations, and two projection-based approaches using the public code for REMBO \cite{wang2013bayesian}, and HeSBO \cite{nayebi2019framework}. In these implementations, the embedding dimension needs to be specified \textit{a priori}, and can have major impacts on performance. We test six different instantiations of REMBO, and HeSBO with different embedding dimensions for the 20, 50, 100, and 1000 dimensional experiments: $d_e = \{6,5,4,3,2,1\}$ for $d=20$, $d_e = \{8,6,4,3,2,1\}$ for $d=50$, $d_e = \{12,10,8,6,2,1\}$ for $d=100$, and $d_e = \{100, 50, 25, 10, 5, 1\}$ for $d=1000$. 
The implementations provided by the authors of the additive Gaussian process based approaches in \cite{wang2017batched} and \cite{rolland2018high} could not scale beyond the 100 dimensional case, and no results were collected in these cases.
    
    \begin{figure}
        \centering
        \subfigure[$d=20$ function value, $\pm2$ standard errors.\label{fig:RepB20val}]{\includegraphics[width=0.45\textwidth, height=0.20\textheight]{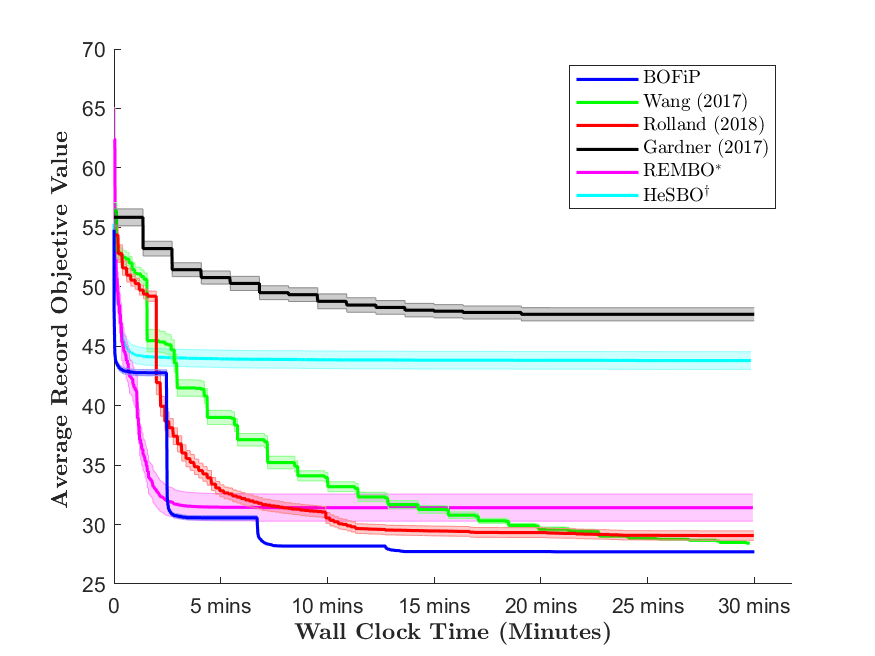}}
        \subfigure[$d=50$ function value, $\pm2$ standard errors.\label{fig:RepB50val}]{\includegraphics[width=0.45\textwidth, height=0.20\textheight]{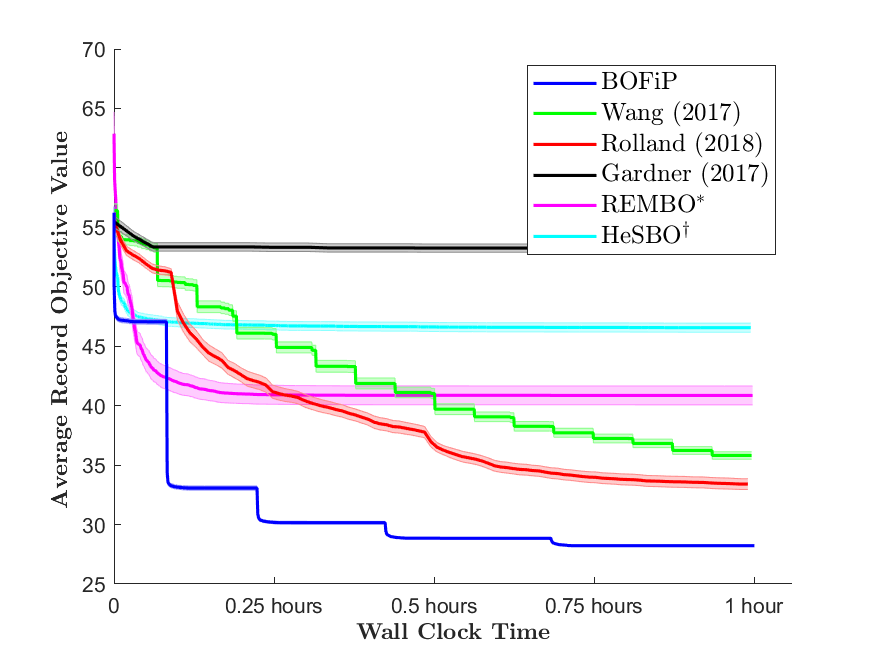}}
        \subfigure[$d=100$ function value, $\pm2$ standard errors.\label{fig:RepB100val}]{\includegraphics[width=0.45\textwidth, height=0.20\textheight]{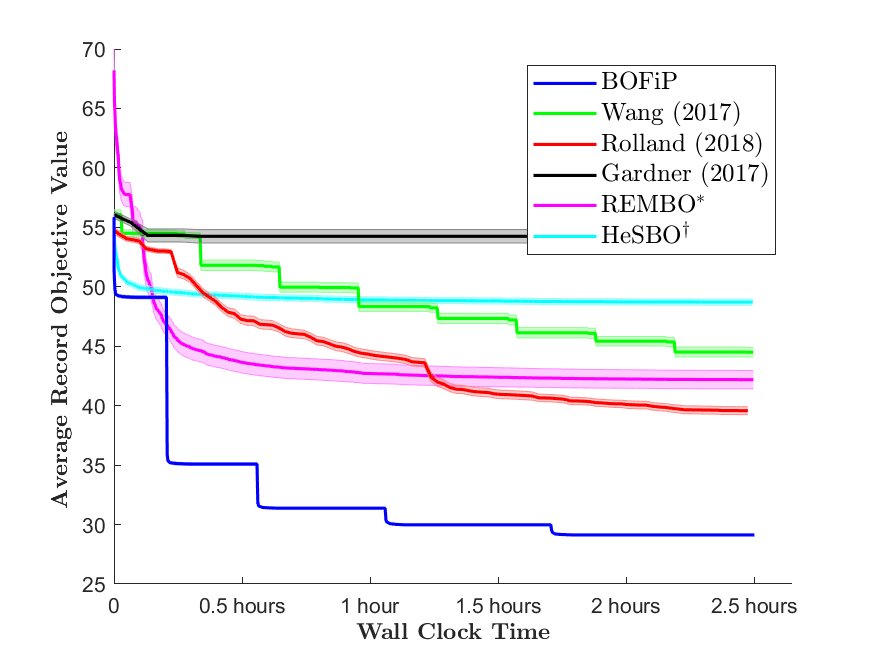}}
        \subfigure[$d=1000$ function value, $\pm2$ standard errors.\label{fig:RepB1000val}]{\includegraphics[width=0.45\textwidth, height=0.20\textheight]{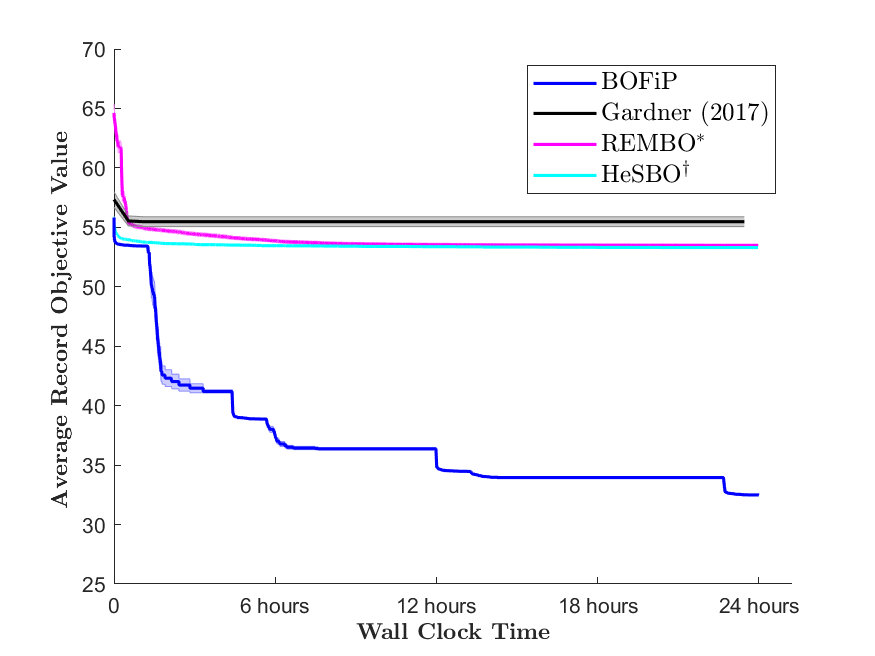}}
        \caption{Repeated Branin function value graphs for 20, 50, 100, and 1000 dimensions.}
        \label{fig:RepBGraphs}
    \end{figure}
    
    \begin{table*}
    \centering
    \caption{Repeated Branin function average best function value optimality gap over 100 replications in 20, 50, 100, and 1000 dimensions, $\pm 2$ standard errors.}
    \label{Tab: RepB Results}
    \resizebox{\textwidth}{!}{%
    \begin{tabular}{ccccccccc}
        \toprule
        &\multicolumn{2}{c}{$d=20$, $d_e = \{6^*,5^{\dag},4,3,2,1\}$} & \multicolumn{2}{c}{$d=50$, $d_e = \{8^{*\dag},6,4,3,2,1\}$} & \multicolumn{2}{c}{$d=100$, $d_e = \{12^{*\dag},10,8,6,2,1\}$} & \multicolumn{2}{c}{$d=1000$, $d_e = \{100,50,25^\dag,10^*,5,1\}$} \\
    	& $\widebar{\lVert\mathbf{x}_i-\mathbf{x}^*\rVert}$ & $\widebar{|f_i-f^*|}$ & $\widebar{\lVert\mathbf{x}_i-\mathbf{x}^*\rVert}$ & $\widebar{|f_i-f^*|}$ & $\widebar{\lVert\mathbf{x}_i-\mathbf{x}^*\rVert}$ & $\widebar{|f_i-f^*|}$ & $\widebar{\lVert\mathbf{x}_i-\mathbf{x}^*\rVert}$ & $\widebar{|f_i-f^*|}$ \\
    	\midrule
        Wang(2017) & $0.47 \pm 0.05$ & $0.74 \pm 0.09$ & $4.29 \pm 0.16$ & $8.11 \pm 0.45$ & $9.23 \pm 0.20$ & $16.79 \pm 0.57$ & N/A & N/A \\
        Rolland(2018) & $1.06 \pm 0.32$ & $1.37 \pm 0.56$ & $4.64 \pm 0.31$ & $5.70 \pm 0.64$ & $9.57 \pm 0.17$ & $11.87 \pm 0.47$ & N/A & N/A \\
        Gardner(2017) & $5.01 \pm 0.14$ & $19.98 \pm 0.78$ & $8.15 \pm 0.19$ & $25.55 \pm 0.49$ & $11.46 \pm 0.26$ & $26.54 \pm 0.80$ & $35.47 \pm 0.56$ & $27.77 \pm 0.59$ \\
        \cline{2-9}     
        REMBO* & $22.17 \pm 0.61$ & $3.72 \pm 1.61$ & $21.53 \pm 0.67$ & $13.16 \pm 1.12$ & $18.00 \pm 0.43$ & $14.49 \pm 1.09$ & $41.21 \pm 0.36$ & $25.76 \pm 0.18$ \\
        HeSBO$^\dag$ & $4.67 \pm 0.17$ & $16.09 \pm 1.05$ & $7.64 \pm 0.105$ & $18.85 \pm 0.54$ & $10.96 \pm 0.14$ & $48.71 \pm 0.37$ & $37.42 \pm 0.13$ & $25.59 \pm 0.11$ \\
        \cline{2-9}
        BOFiP & $\mathbf{0.02 \pm 0.00}$ & $\mathbf{0.00 \pm 0.00}$ & $\mathbf{0.92 \pm 0.12}$ & $\mathbf{0.51 \pm 0.07}$ & $\mathbf{2.74 \pm 0.11}$ & $\mathbf{1.42 \pm 0.07}$ & $\mathbf{15.76 \pm 0.12}$ & $\mathbf{4.78 \pm 0.04}$ \\
    \bottomrule
    \end{tabular}}
    \end{table*}

\noindent\textbf{Analysis.}
Figures~\ref{fig:RepB20val}-\ref{fig:RepB1000val} illustrate the best observed function value, relative to the wall clock time, for the Repeated Branin function in $20, 50, 100,$ and $1000$ dimensions. In the results, REMBO$^*$ and HeSBO$^\dag$ are the implementations of REMBO or HeSBO with the embedding dimension $d_e$ producing the best performance, out of the the six $d_e$ that were tested. 
Observing Figures~\ref{fig:RepB20val}-\ref{fig:RepB1000val}, our proposed BOFiP statistically outperforms all competitors in all tested dimensions. As the dimensionality increases, the performance gap between BOFiP and all competitors widens, showing the scalability of the proposed framework. For all the dimensions (Figure~\ref{fig:RepBGraphs}), we observe how BOFiP's best observed value progresses in a stepwise manner. 
Each step corresponds to a synchronization among sub-spaces. 
Table~\ref{Tab: RepB Results} confirms that BOFiP is the top performer for the Repeated Branin function, additionally, it shows that both REMBO and HeSBO performed better when a larger embedding dimension was used.

   \begin{figure}
        \centering
        \subfigure[$d=20$ function value, $\pm2$ standard errors.\label{fig:RepH620val}]{\includegraphics[width=0.45\textwidth, height=0.20\textheight]{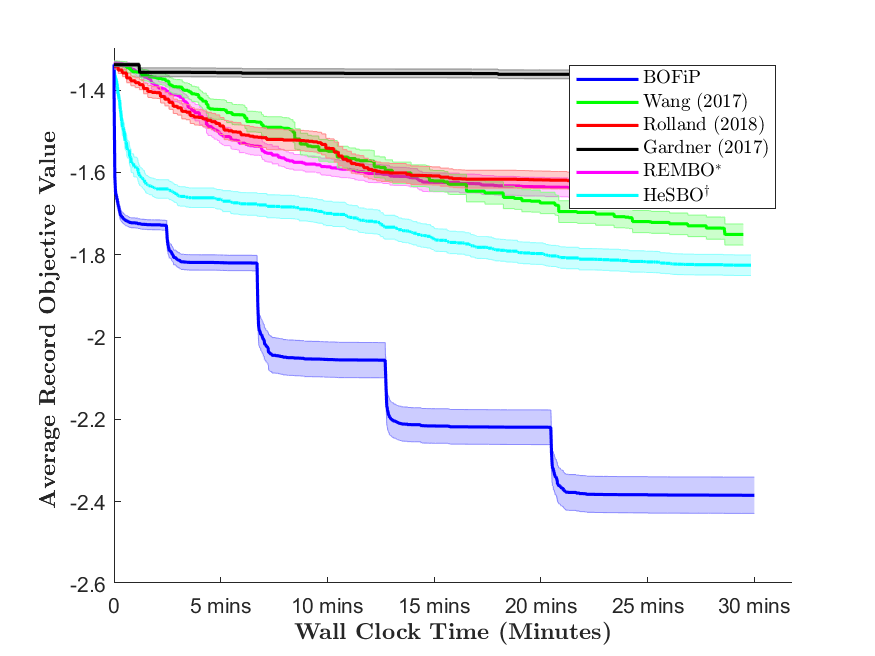}}
        \subfigure[$d=50$ function value, $\pm2$ standard errors.\label{fig:RepH650val}]{\includegraphics[width=0.45\textwidth, height=0.20\textheight]{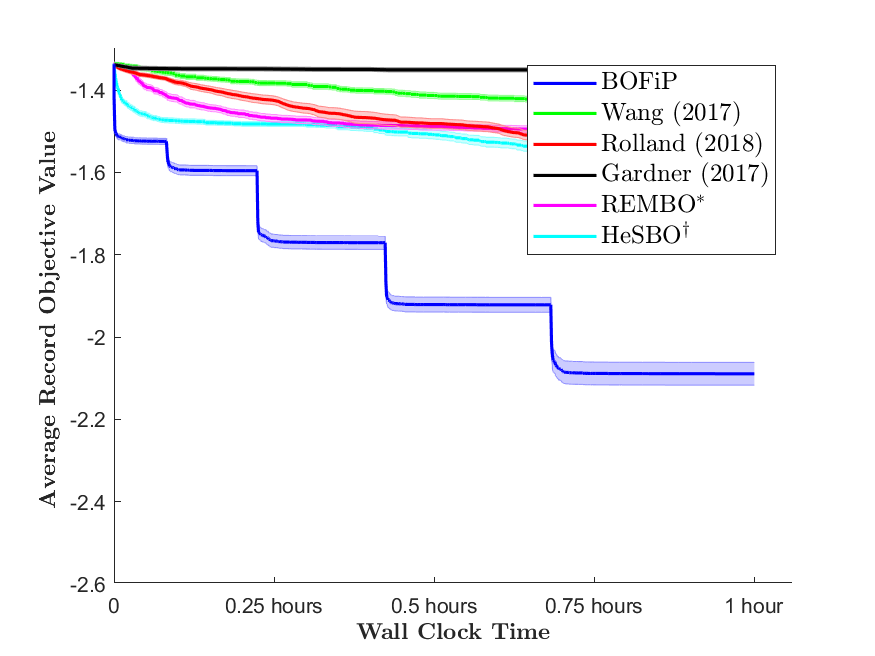}}
        \subfigure[$d=100$ function value, $\pm2$ standard errors.\label{fig:RepH6100val}]{\includegraphics[width=0.45\textwidth, height=0.20\textheight]{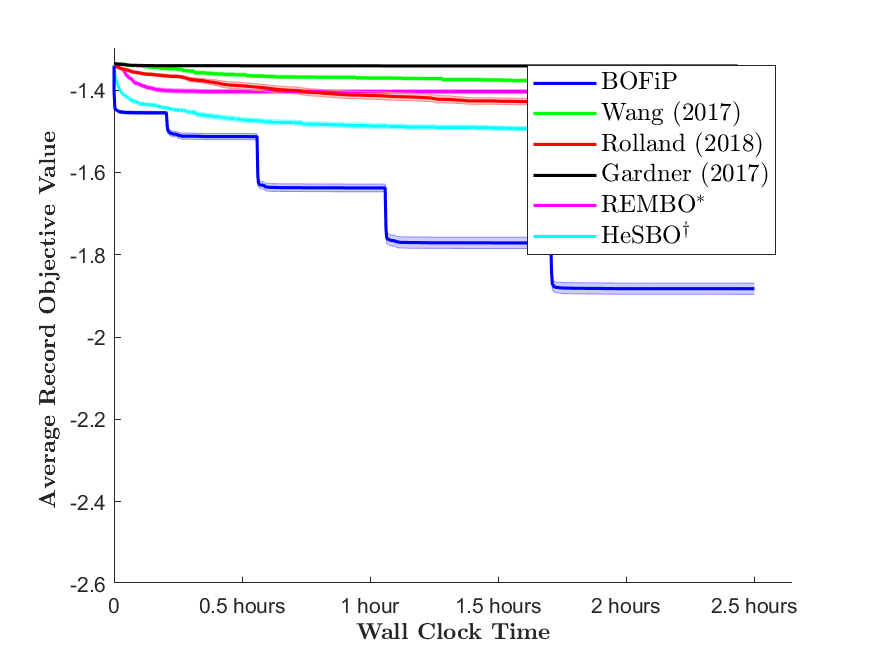}}
        \subfigure[$d=1000$ function value, $\pm2$ standard errors.\label{fig:RepH61000vall}]{\includegraphics[width=0.45\textwidth, height=0.20\textheight]{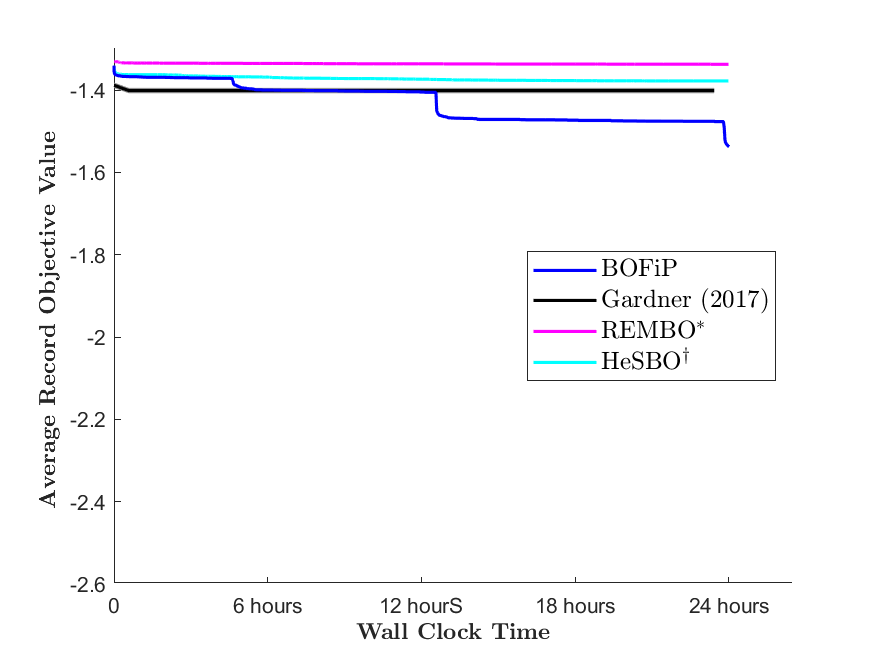}}
        \caption{Repeated Hartmann function value graphs for 20, 50, 100, and 1000 dimensions.}
        \label{fig:RepH6Graphs}
    \end{figure}
   
    \begin{table*}
    \centering
    \caption{Repeated Hartmann function average best function value optimality gap over 100 replications in 20, 50, 100, and 1000 dimensions, $\pm 2$ standard errors.}
    \label{Tab: RepH6 Results}
    \resizebox{\textwidth}{!}{%
    \begin{tabular}{ccccccccc}
        \toprule
        &\multicolumn{2}{c}{$d=20$, $d_e = \{6^{\dag},5,4,3^*,2,1\}$} & \multicolumn{2}{c}{$d=50$, $d_e = \{8,6^{\dag},4,3^*,2,1\}$} & \multicolumn{2}{c}{$d=100$, $d_e = \{12^{\dag},10,8,6,2^*,1\}$} & \multicolumn{2}{c}{$d=1000$, $d_e = \{100,50,25,10^*,5^\dag,1\}$}\\
    	& $\widebar{\lVert\mathbf{x}_i-\mathbf{x}^*\rVert}$ & $\widebar{|f_i-f^*|}$ & $\widebar{\lVert\mathbf{x}_i-\mathbf{x}^*\rVert}$ & $\widebar{|f_i-f^*|}$ & $\widebar{\lVert\mathbf{x}_i-\mathbf{x}^*\rVert}$ & $\widebar{|f_i-f^*|}$ & $\widebar{\lVert\mathbf{x}_i-\mathbf{x}^*\rVert}$ & $\widebar{|f_i-f^*|}$ \\
    	\midrule
        Wang(2017) & $\mathbf{2.78 \pm 0.15}$ & $1.29 \pm 0.04$ & $5.36 \pm 0.17$ & $1.60 \pm 0.01$ & $7.91 \pm 0.14$ & $1.65 \pm 0.00$ & N/A & N/A \\
        Rolland(2018) & $2.17 \pm 0.16$ & $1.42 \pm 0.03$ & $\mathbf{3.30 \pm 0.10}$ & $1.51 \pm 0.01$ & $\mathbf{4.91 \pm 0.18}$ & $1.60 \pm 0.01$ & N/A & N/A \\
        Gardner(2017) & $2.92 \pm 0.15$ & $1.68 \pm 0.01$ & $5.18 \pm 0.20$ & $1.69 \pm 0.00$ & $7.27 \pm 0.24$ & $1.70 \pm 0.00$ & $21.11 \pm 0.49$ & $1.64 \pm 0.01$ \\
        \cline{2-9}
        REMBO* & $2.77 \pm 0.27$ & $1.39  \pm 0.03$ & $5.41 \pm 0.45$ & $1.54 \pm 0.01$ & $6.42 \pm 0.52$ & $1.64 \pm 0.01$ & $67.19 \pm 27.30$ & $1.69 \pm 0.00$\\
        HeSBO$^\dag$ & $\mathbf{2.29 \pm 0.16}$ & $1.21 \pm 0.04$ & $4.01 \pm 0.21$ & $1.47 \pm 0.02$ & $5.49 \pm 0.16$ & $1.55 \pm 0.01$ & $\mathbf{16.56 \pm 0.22}$ & $1.66 \pm 0.00$\\
        \cline{2-9}
        BOFiP & $\mathbf{2.20 \pm 0.13}$ & $\mathbf{0.66 \pm 0.06}$ & $4.14 \pm 0.14$ &  $\mathbf{0.95 \pm 0.04}$ & $6.16 \pm 0.14$ & $\mathbf{1.16 \pm 0.02}$ & $22.83 \pm 0.14$ & $\mathbf{1.50 \pm 0.00}$ \\
    \bottomrule
    \end{tabular}}
    \end{table*}

Figures~\ref{fig:RepH620val}-\ref{fig:RepH61000vall} show the performance of the different algorithms over the Repeated Hartman function. Again, BOFiP has the best performance in all cases, with the record best function value over time exhibiting the same step behavior. Moreover, BOFiP improves at increasingly larger rates (wider steps), whereas all other algorithms show early signs of slowing improvement. This could be related to the increasing iteration cost of the other approaches as compared to BOFiP. It is noteworthy that the Repeated Hartmann function has a large range of variation, and several local minima scattered throughout the domain. 
While the presence of multiple local minima may be challenging to model based approaches, the range of variation may slow down projection based approaches even more if the low effective dimensionality assumption is not satisfied. Table~\ref{Tab: RepH6 Results} shows that both REMBO and HeSBO stabilize around local minima which is near the global minimum.

    \begin{figure}
        \centering
        \subfigure[$d=20$ function value, $\pm2$ standard errors.\label{fig:Ackley20val}]{\includegraphics[width=0.45\textwidth, height=0.20\textheight]{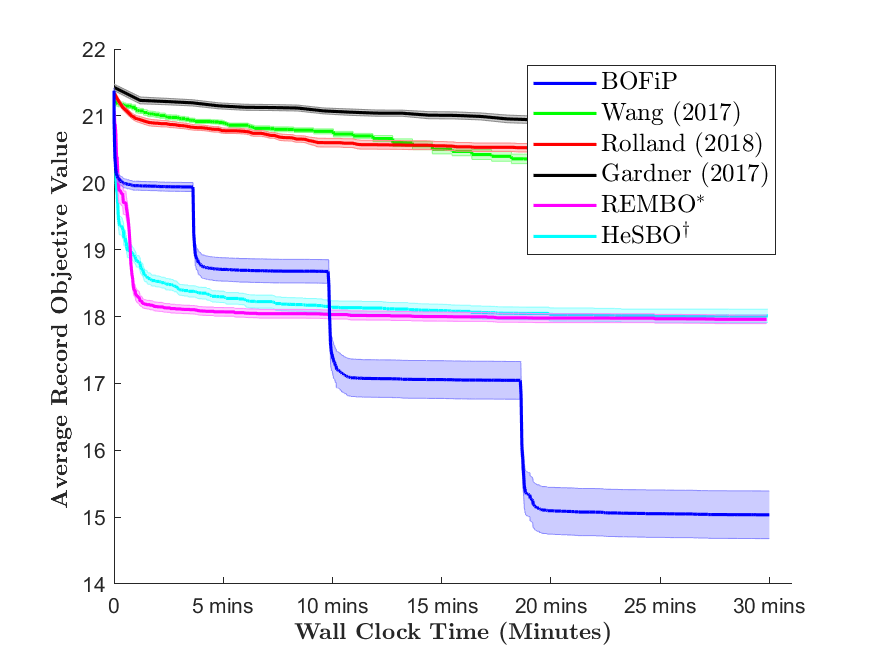}}
        \subfigure[$d=50$ function value, $\pm2$ standard errors.\label{fig:Ackley50val}]{\includegraphics[width=0.45\textwidth, height=0.20\textheight]{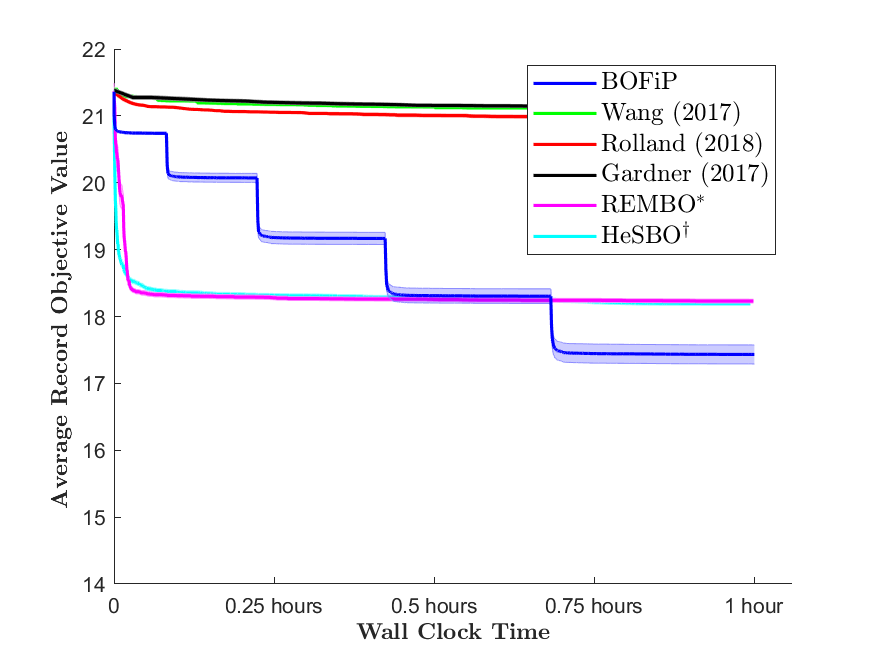}}
        \subfigure[$d=100$ function value, $\pm2$ standard errors.\label{fig:Ackley100val}]{\includegraphics[width=0.45\textwidth, height=0.20\textheight]{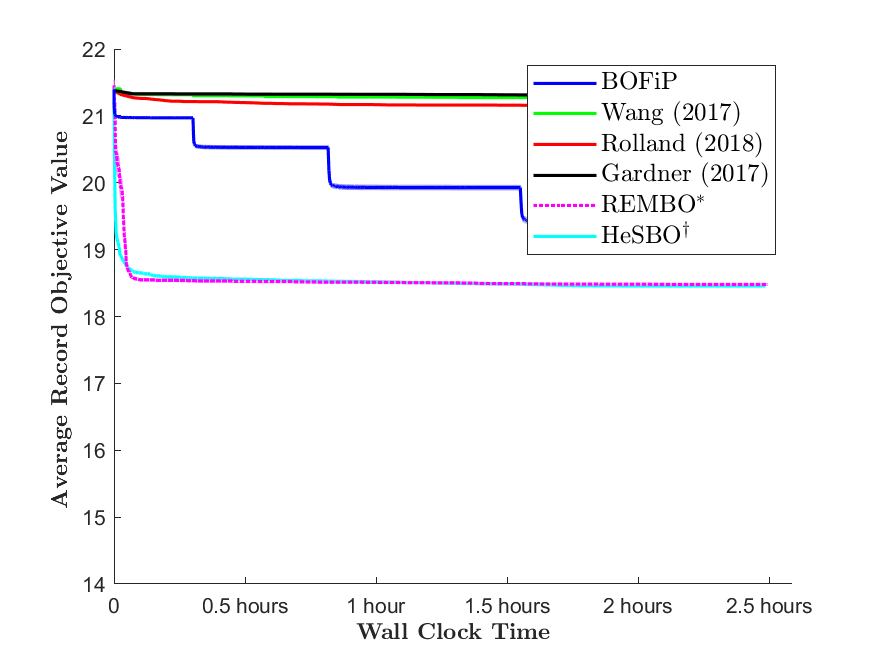}}
        \subfigure[$d=1000$ function value, $\pm2$ standard errors.\label{fig:Ackley1000val}]{\includegraphics[width=0.45\textwidth, height=0.20\textheight]{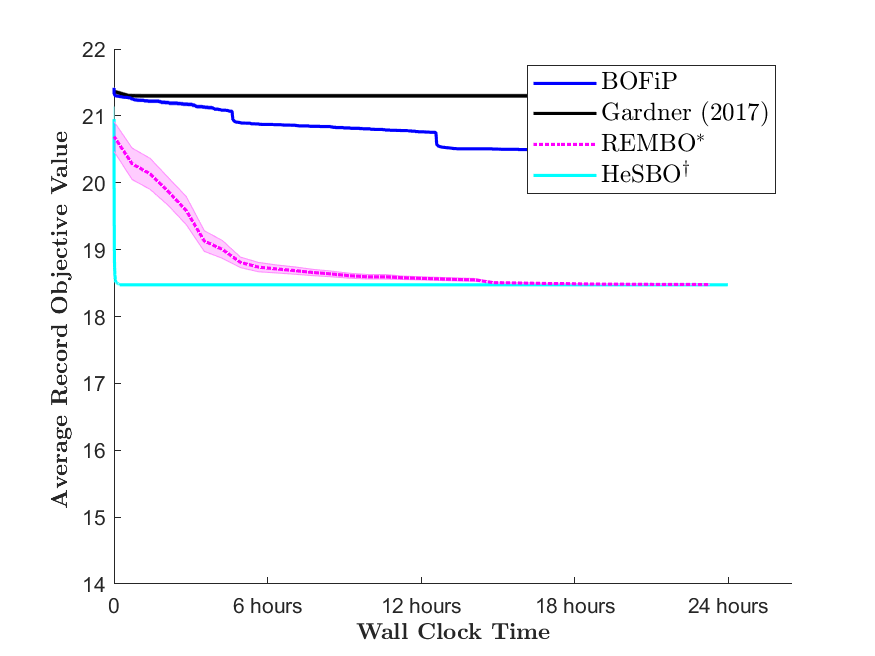}}
        \caption{Ackley function value graphs for 20, 50, 100, and 1000 dimensions.}
        \label{fig:AckleyGraphs}
    \end{figure}


Table~\ref{Tab: Ack Results} reports the best results found by each algorithm over the allowed time window for the Ackley function. Figures~\ref{fig:Ackley20val}-~\ref{fig:Ackley1000val} show algorithm progression over time. 
BOFiP performs statistically better than all additive GP approaches in all cases, and outperforms both embedding approaches in $20$ and $50$ dimensions. 
The highly nonlinear nature of the Ackley function, with thousands of local minima, is again challenging for model based approaches.  
The top embedding approaches project down into one or two dimensions (seen in Table~\ref{Tab: Ack Results}), quickly finding a local minimum solution at the origin.
We can observe how optimizing over the reduced embedded domain allows REMBO and HeSBO to quickly converge to the same objective function value regardless of the dimensionality of the original problem. This can be seen in Figures~\ref{fig:Ackley20val}-\ref{fig:Ackley1000val}, where REMBO and HeSBO always achieve a function value of around $18.5$ with no further improvement. 
Differently, as expected, BOFiP 
is impacted by the problem dimension and the rate of improvement for observed objective function value is slower in the $d=100$ and $d=1000$ cases. However, the steps in BOFiP's performance observed in Figures~\ref{fig:Ackley100val} and~\ref{fig:Ackley1000val} indicate that the algorithm is progressing towards better solutions. 
On the other hand, it is not clear that either REMBO or HeSBO will progress beyond a function value of $18.5$, this indicates that the game theoretic framework that BOFiP proposes is worth studying further. 

\begin{table*}
    \centering
    \caption{Ackley function average best function value optimality gap over 100 replications in 20, 50, 100, and 1000 dimensions, $\pm 2$ standard errors.}
    \label{Tab: Ack Results}
    \resizebox{\textwidth}{!}{%
    \begin{tabular}{ccccccccc}
        \toprule
        &\multicolumn{2}{c}{$d=20$, $d_e = \{6,5,4,3^{\dag},2^*,1\}$} & \multicolumn{2}{c}{$d=50$, $d_e = \{8,6,4,3,2^{*\dag},1\}$} & \multicolumn{2}{c}{$d=100$, $d_e = \{12,10,8,6,2^{*\dag},1\}$} & \multicolumn{2}{c}{$d=1000$, $d_e = \{100,50,25,10,5,1^{*\dag}\}$} \\
    	& $\widebar{\lVert\mathbf{x}_i-\mathbf{x}^*\rVert}$ & $\widebar{|f_i-f^*|}$ & $\widebar{\lVert\mathbf{x}_i-\mathbf{x}^*\rVert}$ & $\widebar{|f_i-f^*|}$ & $\widebar{\lVert\mathbf{x}_i-\mathbf{x}^*\rVert}$ & $\widebar{|f_i-f^*|}$ & $\widebar{\lVert\mathbf{x}_i-\mathbf{x}^*\rVert}$ & $\widebar{|f_i-f^*|}$ \\
    	\midrule
    	Wang(2017) & $63.54 \pm 19.70$ & $20.07 \pm 0.92$ & $139.82 \pm 25.09$ & $21.11 \pm 0.15$ & $204.60 \pm 29.54$ & $21.27 \pm 0.13$ & N/A & N/A \\
    	Rolland(2018) & $67.93 \pm 17.72$ & $20.48 \pm 0.64$ & $126.50 \pm 18.14$ & $20.98 \pm 0.21$ & $188.31 \pm 18.22$ & $21.14 \pm 0.10$ & N/A & N/A\\
    	Gardner(2017) & $77.41 \pm 28.75$ & $20.86 \pm 0.74$ & $134.67 \pm 19.89$ & $21.11 \pm 0.22$ & $206.58 \pm 24.29$ & $21.30 \pm 0.15$ & $620.09 \pm 93.64$ & $21.30 \pm 0.24$ \\
    	\cline{2-9}
    	REMBO* & $39.43 \pm 3.81$ & $17.96 \pm 0.64$ & $63.76 \pm 2.72$ & $18.23 \pm 0.29$  & $93.07 \pm 2.35$ & $18.48 \pm 0.16$ & $288.75 \pm 0.47$ & $18.48 \pm 0.04$ \\
    	HeSBO$^\dag$ & $39.05 \pm 5.79$ & $18.01 \pm 0.99$ & $63.18 \pm 2.57$ & $18.20 \pm 0.30$ & $92.59 \pm 2.19$ & $18.46 \pm 0.17$ & $288.67 \pm 0.31$ & $18.47 \pm 0.01$ \\
    	\cline{2-9}
    	BOFiP & $\mathbf{30.14 \pm 13.99}$ & $\mathbf{15.04 \pm 3.59}$ & $66.32 \pm 14.93$ & $\mathbf{17.43 \pm 1.42}$ & $156.79 \pm 14.81$ & $19.41 \pm 0.44$ & $536.46 \pm 16.25$ & $20.21 \pm 0.09$ \\
    	\bottomrule
    \end{tabular}}
    \end{table*}


    \subsection{Neural Network Design}
    We apply BOFiP for the design of a neural network whose purpose is to accurately classify breast tissue tumors as either cancerous or benign based upon the feature data associated to the tumor.
    A set of 699 patient observations serve as the data set; each observation contains a target variable representing a cancer classification and 8 tumor feature variables~\cite{murphy1994uci}. 
    Three NN architecture design experiments are completed over this dataset. Fully connected feed forward NN architectures are initially specified with a maximum of: 502, 1012, or 10002 individual weight components; if a weight component is set to zero it effectively removes the associated node from the NN architecture. The three architectures have 6, 11, and 92 layers respectively, with a maximum of 10 nodes per layer. The goal of the test is for the black box optimization algorithm to determine the value for each of the 
    weight components to 
    minimize the observered MSE when the NN is used to classify the 699 tumor observations. 
    
    We compare BOFiP against REMBO, considering $6$ different embedding dimensions for each experiment: $d_e = \{50,25,10,5,2,1\}$ for $d=502$, $d_e = \{100,50,25,10,5,1\}$ for $d=1012$, and $d_e = \{500,100,50,25,10,1\}$ for $d=10002$. We set a wall-clock time limit for all tests, allowing 12 hours of computation time for $d=502$, 24 hours for $d=1012$, and 24 hours for $d=10002$. 
    \begin{figure}[t]
        \centering
        \subfigure[$d=502$ function value, $\pm2$ standard errors.\label{fig:NN500val}]{\includegraphics[width=0.45\textwidth, height=0.20\textheight]{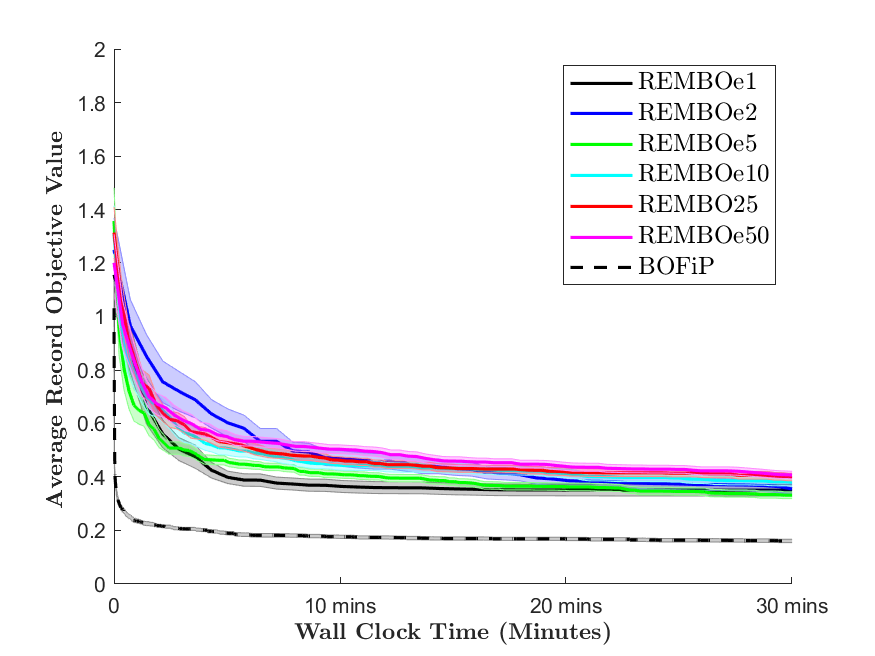}}
        \subfigure[$d=1012$ function value, $\pm2$ standard errors.\label{fig:NN1012val}]{\includegraphics[width=0.45\textwidth, height=0.20\textheight]{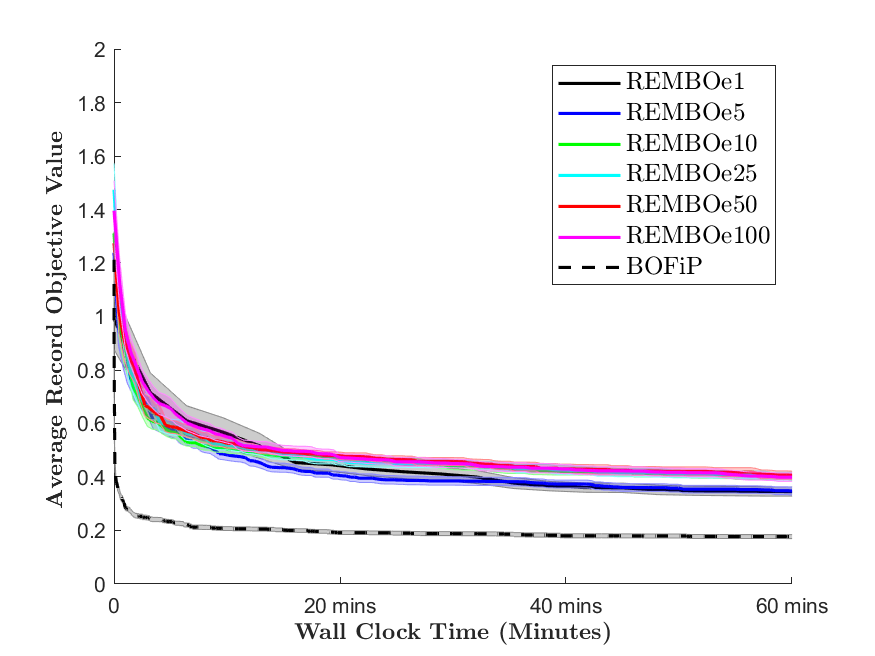}}
        \subfigure[$d=10002$ function value, $\pm2$ standard errors.\label{fig:NN10002val}]{\includegraphics[width=0.45\textwidth, height=0.20\textheight]{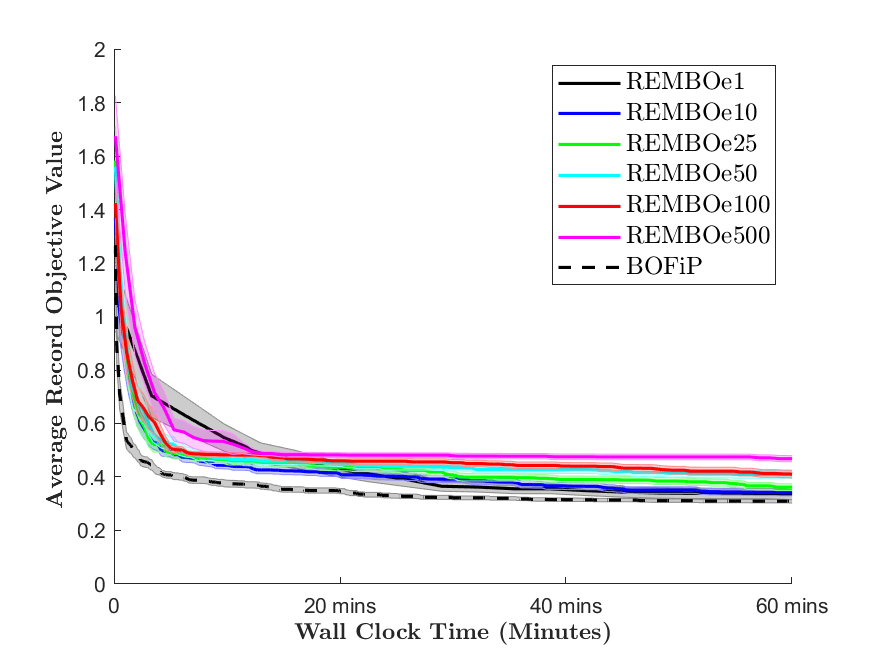}}
        \caption{Neural Net weight fitting mean square error value graphs for 502, 1012, and 10002 dimensional cases.}
        \label{fig:NNGraphs}
    \end{figure}
    
    \begin{table*}
    \centering
    \caption{Neural Network best observed mean square error over 100 replications in 502, and 1012 dimensional testing, $\pm 2$ standard errors.}
    \label{Tab: NN Results}
    \resizebox{0.75\textwidth}{!}{%
    \begin{tabular}{ccccccc}
        \toprule
    	& $d=502$ & $\widebar{|f_i|}$ & $d=1012$ & $\widebar{|f_i|}$ & $d=10002$ & $\widebar{|f_i|}$ \\
    	\midrule
    	\multirow{6}{*}{\rotatebox[origin=c]{90}{REMBO}} 
    	& $d_e = 1$ & $0.3506 \pm 0.0309$ & $d_e = 1$ & $0.3227 \pm 0.0258$ & $d_e = 1$ & $0.2974 \pm 0.0225$\\
    	& $d_e = 2$ & $0.2002 \pm 0.0184$ & $d_e = 5$ & $0.2149 \pm 0.0078$ & $d_e = 10$ & $0.2545 \pm 0.0059$\\
    	& $d_e = 5$ & $0.1886 \pm 0.0081$ & $d_e = 10$ & $0.2397 \pm 0.0100$ & $d_e = 25$ & $0.2693 \pm 0.0042$\\
    	& $d_e = 10$ & $0.2172 \pm 0.0090$ & $d_e = 25$ & $0.2468 \pm 0.0088$ & $d_e = 50$ & $0.2775 \pm 0.0058$\\
    	& $d_e = 25$ & $0.2405 \pm 0.0089$ & $d_e = 50$ & $0.2444 \pm 0.0101$ & $d_e = 100$ & $0.2728 \pm 0.0075$\\
    	& $d_e = 50$ & $0.2482 \pm 0.0091$ & $d_e = 100$ & $0.2591 \pm 0.0090$ & $d_e = 500$ & $0.3102 \pm 0.0107$\\
    	\cline{1-7}
    	& BOFiP & $\mathbf{0.1335 \pm 0.0080}$ & BOFiP & $\mathbf{0.1278 \pm 0.0080}$ & BOFiP & $\mathbf{0.2507 \pm 0.0210}$\\
      	\bottomrule
    \end{tabular}}
    \end{table*}
    
    The results 
    are shown in Table~\ref{Tab: NN Results}, and in Figure~\ref{fig:NNGraphs}. Table~\ref{Tab: NN Results} shows that BOFiP outperforms REMBO under any tested embedding dimension. 
    Figure~\ref{fig:NNGraphs} illustrates the performance as a function of the wall clock time (only part is shown since the remainder of the experiment showed minimal change in performance for all of the algorithms, and it was omitted). Across the three test cases BOFiP quickly and consistently identifies the architecture 
    that produces the best training results, and the embedding dimension appears to play only a marginal role for REMBO. 
    It is noteworthy how the decomposition idea underlying BOFiP appears to be well suited for problems such as NN architecture design. While in our implementation each weight in the NN is a player, implementations where each layer is a player could also be considered.
    
    
    
    
    

\section{Conclusion}
BOFiP scales 
Bayesian optimization through the use of sampled fictitious play, which solves for the equilibria of equal interest games. 
We show how a high dimensional optimization problem can be decomposed into sub-problems with associated sub-spaces and objective functions, approximately solved using a Bayesian optimizer.
These sub-spaces are players in an equal interest game, and belief vectors are the mechanism to discover the location of the optimal solution. 

The interplay of the game theoretic fictitious play mechanism and Bayesian optimization is the first to the authors knowledge. Also, we provide a new implementation of sampled fictitious play where exact best replies are not required. Moreover, BOFiP applies without making assumptions on the problem being solved, i.e., additive model structure or low effective dimensionality. Finally, the BOFiP framework can be integrated with other high dimensional approaches, replacing the Bayesian optimization sub-space optimizer with alternative algorithms. 

Our numerical results show that BOFiP finds good solutions in a short amount of time, relative to other state-of-the-art high dimensional Bayesian optimization methods, and show BOFiP's ability to computationally scale to very high dimensional spaces. Ongoing and future work include: \begin{inparaenum}
\item[(1)] adaptive and dynamic dimensionality assignment to sub-spaces, with overlapping dimensionality among sub-spaces, \item[(2)] adaptive budget assignment for Bayesian optimization, \item[(3)] use of new fictitious play algorithms, \item[(4)] Theoretical analysis of the algorithm performance.
\end{inparaenum}


%
%
%


\bibliography{ref_bosfp}

\end{document}